\documentclass[11pt]{article}

\usepackage{PRIMEarxiv}
\usepackage[utf8]{inputenc} % allow utf-8 input
\usepackage[T1]{fontenc}    % use 8-bit T1 fonts
\usepackage{lineno}
\usepackage{booktabs}% http://ctan.org/pkg/booktabs
\usepackage{array}% http://ctan.org/pkg/array
\usepackage{tabularx}% http://ctan.org/pkg/tabularx
\usepackage{bigstrut}
\usepackage{amssymb}
\usepackage{caption}
\usepackage{subcaption}
\usepackage{algorithm}
\usepackage{algorithmic}
\usepackage{amsmath}
\usepackage{graphicx}
\usepackage{multirow}
\usepackage{multicol}
\usepackage{graphicx}
\usepackage{multirow}
\usepackage{graphicx}
\usepackage[table,xcdraw]{xcolor}
\usepackage{orcidlink}
\usepackage{hhline}
\usepackage[normalem]{ulem}
\usepackage{caption}
\usepackage{graphicx}
\usepackage{multirow}
\usepackage{colortbl}
\usepackage{booktabs}
\usepackage{tablefootnote}

\title{VINet: Lightweight, Scalable, and Heterogeneous Cooperative Perception for 3D Object Detection
}

\author{
  Zhengwei Bai\thanks{\textit{Corresponding Author}}, Guoyuan Wu, Matthew J. Barth \\
  University of California, Riverside \\
  Riverside, CA\\
  \texttt{zbai012@ucr.edu, gywu@cert.ucr.edu, barth@ece.ucr.edu} \\
   \And
  Yongkang Liu, Emrah~Akin~Sisbot, Kentaro Oguchi \\
  Toyota North America InfoTech Labs, \\
  Mountain View, CA\\
  \texttt{\{yongkang.liu, akin.sisbot,  kentaro.oguchi\}@toyota.com}
}

\begin{document}
\maketitle

\begin{abstract}
Utilizing the latest advances in Artificial Intelligence (AI), the computer vision community is now witnessing an unprecedented evolution in all kinds of perception tasks, particularly in object detection. Based on multiple spatially separated perception nodes, Cooperative Perception (CP) has emerged to significantly advance the perception of automated driving. However, current cooperative object detection methods mainly focus on ego-vehicle efficiency without considering the practical issues of system-wide costs. In this paper, we introduce VINet, a unified deep learning-based CP network for scalable, lightweight, and heterogeneous cooperative 3D object detection. VINet is the first CP method designed from the standpoint of large-scale system-level implementation and can be divided into three main phases: 1) Global Pre-Processing and Lightweight Feature Extraction which prepare the data into global style and extract features for cooperation in a lightweight manner; 2) Two-Stream Fusion which fuses the features from scalable and heterogeneous perception nodes; and 3) Central Feature Backbone and 3D Detection Head which further process the fused features and generate cooperative detection results. An open-source data experimental platform is designed and developed for CP dataset acquisition and model evaluation. The experimental analysis shows that VINet can reduce 84\% system-level computational cost and 94\% system-level communication cost while improving the 3D detection accuracy.
\end{abstract}

% keywords can be removed

%% main text
\section{Introduction}
\label{intro}
%% \linenumbers
Throughout the world, transportation demand has increased in terms of the movement of people and goods on a daily basis. It is important to note, however, that a rapidly increasing number of vehicles has resulted in several major problems to the current transportation system, including safety~\cite{2019Crash}, mobility~\cite{2018Congestion}, and environmental sustainability~\cite{2021Energy}. In order to improve system-wide performance, Cooperative Driving Automation (CDA) has emerged, leveraging recent advances in advanced sensing, wireless connectivity, and artificial intelligence. CDA enables connected and automated vehicles (CAVs) to communicate with one another, with the  roadway infrastructure, and/or with pedestrians and cyclists equipped with mobile devices. In the past few years, CDA has gained increasing attention and is now considered a transformative solution to these challenges~\cite{fagnant2015preparing}. CDA applications are fundamentally dependent on object perception data from the surrounding environments, which is similar to the visual function of automated agents~\cite{2021SAE}. As the system input, perception data can support a variety of CDA applications, such as Collision Warning~\cite{wu2020improved}, Eco-Approach and Departure (EAD)~\cite{bai2022hybrid}, and Cooperative Adaptive Cruise Control (CACC)~\cite{wangCACC}.

\begin{figure}[!t]
    \centering
    \includegraphics[width=\textwidth]{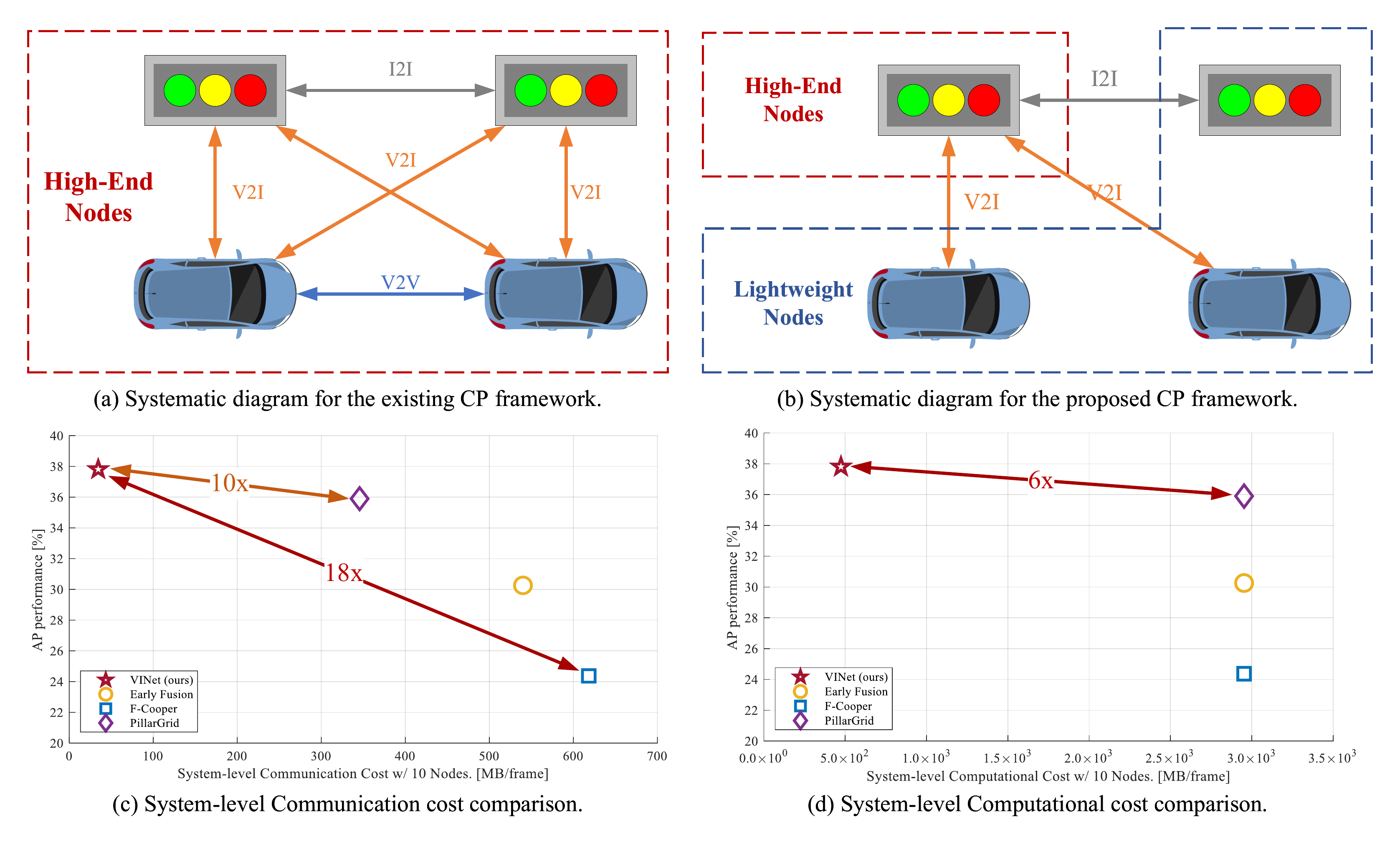}
    \caption{\textbf{The proposed framework for scalable, lightweight, and heterogeneous CP.} For a certain area, a central node is equipped with powerful computing resources while other nodes are only equipped with lightweight computing resources. Lightweight nodes will generate features for cooperation with limited computing power and the central node will generate the cooperative perception results and then broadcast to all road users (i.e., vehi. and infra.).}
    \label{fig: concept}
\end{figure}

% \textit{4) Introduce object detection in terms of sensor types.}\\

In the past few decades, sensing technology has made it possible for transportation systems to retrieve high-fidelity traffic data from various types of sensors. The ability of situation awareness varies among different sensors both onboard vehicles or at the roadside. Cameras can provide detailed vision data for classifying various kinds of traffic objects, including vehicles, pedestrians, and cyclists~\cite{liu2020deep}. On the other hand, high-fidelity 3D point cloud data can be retrieved by LiDAR sensors to grasp the precise 3D location of the traffic objects~\cite{arnold2019survey}. The robust performance of radar sensors in variable environmental conditions has made them an integral part of safety-critical automotive applications~\cite{8443497}.

Perceiving the surrounding environment based on intelligent entities themselves has been one of the main methodologies to promote the development of Autonomous Driving Technology (ADT)~\cite{yurtsever2020survey, arnold2019survey}. For automakers, automated vehicles are now equipped with more onboard sensors and powerful mobile computing that can process the massive amount of sensor data~\cite{waymoADS, lindholm2008nvidia}. Not only for onboard sensors, but recent researchers are now also conducting some studies that use infrastructure sensors that have the capability to perceive the traffic conditions at the object level~\cite{bai2022cmm, zhao2019detection}. These different sensing entities can be regarded as spatially separated perception nodes (PNs) in a transportation system. Even when empowered with advanced perception methods, both Vehicle-based PNs (V-PNs) and Infrastructure-based PNs (I-PNs) have their specific and common limitations to perceiving the environment on their own. Specifically, V-PNs are inevitably limited by the occlusion of other road objects, while I-PNs are limited by the field of view (FOV) and blind zones. From a common point of view, neither V-PNs nor I-PNs themselves can perceive objects that are physically occluded or out of their sensing range. Thus to get past the bottleneck of single-PN perception, cooperation perception has emerged and is attracting increasing attention from researchers.

From an intuitive perspective, collaborating spatially separated PNs, i.e., Cooperative Perception (CP), naturally becomes a transformative solution to improving the perception breadth and accuracy. Therefore, numerous studies have been conducted to generate the perception information in a cooperative manner~\cite{arnold2020cooperative, F-cooper, bai2022pillargrid, wang2020v2vnet, xu2022v2x}. In terms of the stage of sensor fusion, CP can be divided into three main categories: 1) early fusion in which raw sensor data is shared and combined~\cite{chen2019cooper}, 2) late fusion in which perception results (i.e., bounding boxes) from each PNs are fused~\cite{arnold2020cooperative}, and 3) deep fusion in which feature data from each PNs are shared and fused~\cite{bai2022pillargrid}. Different fusion schemes have their own merits and drawbacks. Early fusion only needs the calibration for aligning multi-source data into a unified coordinate system but requires a large communication bandwidth for transmitting data and is more sensitive to noise and delay~\cite{xu2022v2x} and is difficult to fuse multi-modal sensor data. Late fusion mainly focuses on how to merge the perception results generated from multiple perception pipelines, which is straightforward but suffers from limited accuracy~\cite{arnold2020cooperative}. Deep fusion is capable of generating high-performance perception results with low-communication requirements (compared to early fusion), but the development is still in its infancy and only a few studies have been  conducted~\cite{F-cooper, wang2020v2vnet}. 

Furthermore, due to the heterogeneity of V-PNs and I-PNs, only a couple of studies have focused on the vehicle-infrastructure CP~\cite{bai2022pillargrid, xu2022v2x}. A real-world traffic scenario is always a dynamically changing system, which requires the CP methods to be able to cope with scalable PNs. Meanwhile, from the perspective of commercial deployment, it could be extremely costly to enable all PNs with powerful mobile computing units. However, to the best of our knowledge, the CP method that considers all the aspects mentioned above is still missing.

Hence, in this paper, we propose a cooperative perception network named \textit{VINet} (\textit{\textbf{V}ehicle-\textbf{I}nfrastructure \textbf{Net}work}), a scalable and lightweight CP network that supports the cooperation of heterogeneous PNs. VINet mainly consists of five components: 1) Global Pre-Processing (GPP) which transforms the raw sensor data based on the global referencing coordinate; 2)Lightweight Feature Extraction (LFE) which generates deep features from raw sensor data with low-computational requirements; 3) Two-Stream Fusion (TSF) that fuses the features generated from scalable and heterogeneous PNs; 4) Central Feature Backbone (CFB) that extract hidden feature from the fused data; and 5) 3D Detection Head (3DH) that generates rotated 3D bounding boxes. The main contributions of this paper can be summarized as follows:
\begin{itemize}
    \item We introduce the first unified CP framework to offer object detection with low system-level communication and computational cost;
    \item We propose VINet, the first deep learning-based scalable, lightweight, and heterogeneous CP method;
    \item We propose TSF, a novel deep-fusion model for heterogeneous feature data from scalable PNs; and
    \item We design and develop an open-source cooperative perception platform for supporting CP training and evaluation in scalable and heterogeneous environments.
    % \item Numerous experiments have been conducted to assess the proposed method with several baselines.
\end{itemize}

The remainder of this paper is organized as follows: related work is reviewed in Section~\ref{sec: overview} to give a quick glance at the basic CP background. VINet is introduced in Section~\ref{sec: VINet} with details of each consisting component. Section~\ref{sec: experiments} illustrates experimental details about the training and evaluation process, followed by Section~\ref{sec: conclusion} that concludes the paper and highlights future work.

\section{RELATED WORK}
\label{sec: overview}
\subsection{General 3D Object Detection From Point Clouds}
Before 2015, one of the most popular mechanisms to handle point cloud data (PCD) from LiDAR sensors is the bottom-up pipeline based on traditional methods, such as ``\textit{Clustering \cite{ahmed2020density}$\rightarrow$Classification \cite{zhang2016multilevel}$\rightarrow$Tracking \cite{welch1995introduction}}''. Due to their explainability and free from data labeling, traditional bottom-up methodologies are still popular in current infrastructure-based LiDAR perception tasks~\cite{zhang2019vehicle,zhao2019detection,zhang2019automatic}. 

With the great success achieved by convolutional neural networks (CNNs) in image-based object perception, PCD quickly became the upcoming target for CNNs. Point-wise manipulation is considered straightforward for extracting features from PCD for object detection~\cite{qi2017pointnet++, shi2019pointrcnn}. Endowed with the natural fit, point-based methods provide dominant performance in detection accuracy, however, it scarifies computational efficiencies~\cite{shi2020pv}. 

Since PCD consists of 3-dimensional sparse data, 
% \cite{wang2015voting} 
creatively cut the whole 3D point cloud into 3D voxels grids. A specific voting scheme, named \textit{Vote3D}, was designed in 2015. In 2018, VoxelNet~\cite{zhou2018voxelnet} was proposed, which introduced a learnable voxel encoder to generate hidden features of voxels. This voxelization mechanism has been widely used in various works, such as \textit{SECOND}~\cite{yan2018second}, \textit{PointPillar}~\cite{lang2019pointpillars}, \textit{Voxel RCNN}
% ~\cite{deng2021voxel}, etc. 

Projecting PCD into a 2D bird's-eye view (BEV) feature map has quickly become a popular methodology. Inspired by YOLO, \cite{simony2018complex} proposed \textit{ComplexYolo} which projected PCD into three manually defined feature channels, and then the BEV feature map was fed into a 2D backbone for generating detection results. Since the BEV scheme provides a straightforward way for solving 3D data in 2D manners, lots of BEV-based methods have emerged such as \textit{PIXOR}~\cite{yang2018pixor}, \textit{SCANet}~\cite{lu2019scanet}, \textit{BEVFusion}~\cite{liu2022bevfusion}, etc.

\subsection{Roadside LiDAR-based Object Detection}
In recent years, roadside LiDAR sensors have received increasing attention from researchers about object perception in transportation~\cite{bai2022infrastructure}. Using roadside LiDAR, Zhao~et~al. proposed a detection and tracking approach for pedestrians and vehicles~\cite{zhao2019detection}. As one of the early studies utilizing roadside LiDAR for perception, a classical detection and tracking pipeline for PCD was designed. It mainly consists of 1) \textit{Background Filtering}: To remove the laser points reflected from road surfaces or buildings by applying a statistics-based background filtering method~\cite{wu2017automatic}; 2) \textit{Clustering}: To generate clusters for the laser points by implementing a Density-based spatial clustering of applications (DBSCAN) method
% ~\cite{ester1996density};
3) \textit{Classification}: To generate different labels for different traffic objects, such as vehicles and pedestrians, based on neural networks~\cite{li2012brief}; and 4) \textit{Tracking}: To identify the same object in continuous data frames by applying a discrete Kalman filter~\cite{welch1995introduction}.

Based on the aforementioned work, Cui~et~al. designed an automatic vehicle tracking system by considering vehicle detection and lane identification~\cite{cui2019automatic}. In addition, a real-world operational system was developed, consisting of a roadside LiDAR, an edge computer, a \textit{Dedicated Short-Range Communication} (\textit{DSRC}) \textit{Roadside Unit} (\textit{RSU}), a Wi-Fi router, and a DSRC \textit{On-board Unit} (\textit{OBU}), and a GUI. Following a similar workflow, Zhang~et~al. proposed a vehicle tracking and speed estimation approach based on a roadside LiDAR~\cite{zhang2020vehicle}. Vehicle detection results were generated by the ``\textit{Background Filtering-Clustering-Classification}'' process. Then, a centroid-based tracking flow was implemented to obtain initial vehicle transformations, and the unscented Kalman Filter~\cite{julier2004unscented} and joint probabilistic data association filter~\cite{bar2009probabilistic} were adopted in the tracking flow. Finally, vehicle tracking was refined through a BEV LiDAR-image matching process to improve the accuracy of estimated vehicle speeds. Following the bottom-up pipeline mentioned above, numerous roadside LiDAR-based methods have been proposed from various points of view~\cite{zhang2020gc, Song9216093, gouda2021automated, 8484040, zhang2019automatic}. 

On the other hand, using learning-based models to cope with LiDAR data is another primary stream. Bai~et~al.~\cite{bai2022cyber} proposed a deep-learning-based real-time vehicle detection and reconstruction system from roadside LiDAR data. Specifically, the CARLA simulator~\cite{dosovitskiy2017carla} was implemented for collecting the training dataset, and ComplexYOLO model~\cite{simony2018complex} was applied and retrained for object detection. Finally, a co-simulation platform was designed and developed to provide vehicle detection and object-level reconstruction, which aimed to empower subsequent cooperative driving automation (CDA) applications with readily retrieved authentic detection data. In their following work for real-world implementation, Bai~et~al.~\cite{bai2022cmm} proposed a deep-learning-based 3D object detection, tracking, and reconstruction system for real-world implementation. The field operational system consisted of three main parts: 1) 3D object detection by adopting PointPillar~\cite{lang2019pointpillars} for inference from roadside PCD; 2) 3D multi-object tracking by improving DeepSORT~\cite{veeramani2018deepsort} to support 3D tracking, and 3) 3D reconstruction by geodetic transformation and real-time onboard \textit{Graphic User Interface} (\textit{GUI}) display. 

By combining traditional and deep learning algorithms Gong~et~al. \cite{gong2021pedestrian} proposed a roadside LiDAR-based real-time detection approach. Several techniques were designed to guarantee real-time performance, including the application of Octree with region-of-interest (ROI) selection, and the development of an improved Euclidean clustering algorithm with an adaptive search radius. The roadside system was equipped with \textit{NVIDIA Jetson AGX Xavier}, achieving the inference time of $110 ms$ per frame.

\subsection{Multi-Node Cooperative Object Detection}
Although one single LiDAR can provide panoramic FOV around the ego-vehicle, physical occlusion may easily block the lines of sight and cause the ego-vehicle to lose some crucial perception information, which significantly affects its decision-making or control process. Additionally, a spatially separated LiDAR perception system can expand the perceptive range for intelligent vehicles or smart infrastructure.

One of the straightforward inspirations of the multi-LiDAR perception system is sharing the raw PCD via V2V communications~\cite{chen2019cooper}. However, limited wireless communication bandwidth may significantly limit real-time performance. Feature data generated from CNN requires much less bandwidth and is more robust to sensor noise, thus becoming a popular solution to multi-LiDAR fusion~\cite{F-cooper,wang2020v2vnet}. Marvasti~et~al.~\cite{marvasti2020cooperative} used two sharing-parameter CNNs to extract the feature map for PCD retrieved from two-vehicle nodes. Feature maps were then aligned based on the relative position and fused by element-wise summation. By applying an attention mechanism, Xu~et~al.~\cite{xu2021opv2v} proposed a V2V-based cooperative object detection method. A similar CNN process~\cite{lang2019pointpillars} was designed for extracting feature maps for V2V sharing. Furthermore, self-attention was involved in data aggregation based on spatial location in the feature map.

Recently, researchers started focusing on cooperation between V-PN and I-PN based on the multi-LiDAR system. For handling the data heterogeneity from roadside and onboard PCD, Bai~et~al.~\cite{bai2022pillargrid} proposed a decoupled multi-stream CNN framework for generating feature maps accordingly. Relative position information was applied to PCD alignment and the shared feature maps were then fused based on grid-wise \textit{maxout} operation. Additionally, Xu~et~al.~\cite{xu2022v2x} proposed a ViT-based CP method for heterogeneous PNs. Feature maps were extracted using sharing-parameter CNNs and V2X communications. For dealing with heterogeneity, specific graph transformer structures were designed for data extraction.

\begin{table*}[!t]
  \centering
  \caption{Summary of Different Fusion Schemes for Cooperative Perception \cite{bai2022survey}.}
  \resizebox{\textwidth}{!}{%
    \begin{tabular}{c|c|p{25em}|c|c}
    \toprule
    \multicolumn{1}{c|}{Fusion Scheme} & 
    \multicolumn{1}{c|}{Methodology} & 
    \multicolumn{1}{c|}{Pros. and Cons.} & 
    \multicolumn{1}{c|}{Highlighted Features} & 
    \multicolumn{1}{c}{Instance} \\
    \midrule
    \multicolumn{1}{c|}{\multirow{2}[4]{*}{Early Fusion}} & \multicolumn{1}{c|}{\multirow{2}[4]{*}{Deep Learning}} & Pros: Raw data is shared and gathered to form a holistic view.  & \multicolumn{1}{c|}{\multirow{2}[4]{13em}{Raw point cloud data is compressed to fit the limited bandwidth.}} & \multicolumn{1}{c}{\multirow{2}[4]{*}{Chen~et~al.~\cite{chen2019cooper}}} \\
\cmidrule{3-3}      &   & Cons: Low tolerance to the noise and delay of the transmitted data; potentially constrained by the communication bandwidth. &   &  \\
    \midrule
    \multicolumn{1}{c|}{\multirow{2}[4]{*}{Deep Fusion}} & \multicolumn{1}{c|}{\multirow{2}[4]{*}{Deep Learning}} & Pros: High tolerance to the noise, delay, and difference between different nodes and sensor models. & \multicolumn{1}{c|}{\multirow{2}[4]{13em}{Deep neural features are extracted and fused based on spatial correspondence.}} & \multicolumn{1}{c}{\multirow{2}[4]{*}{Bai~et~al.~\cite{bai2022pillargrid}}} \\
\cmidrule{3-3}      &   & Cons: Require training data and hard to find a systematic way for model design. &   &  \\
    \midrule
    \multicolumn{1}{c|}{\multirow{2}[4]{*}{Late Fusion}} & \multicolumn{1}{c|}{\multirow{2}[4]{*}{Traditional}} & Pros: Easy to design and deploy in a real-world system. & \multicolumn{1}{c|}{\multirow{2}[4]{13em}{A late-fusion is proposed based on joint re-scoring and non-maximum suppression.}} & \multicolumn{1}{c}{\multirow{2}[4]{*}{Zhang~et~al.~\cite{zhang2021distributed} }} \\
\cmidrule{3-3}      &   & Cons: Significantly limited by the wrong perception results or the difference between sources. &   &  \\
    \bottomrule
    \end{tabular}}%
  \label{tab: fusion}%
\end{table*}%

\section{Methodology}
\label{sec: VINet}
% In this section, the VINet structure is introduced and each module of VINet is illustrated in detail.

\begin{figure*}[!ht]
    \centering
    \includegraphics[width=\textwidth]{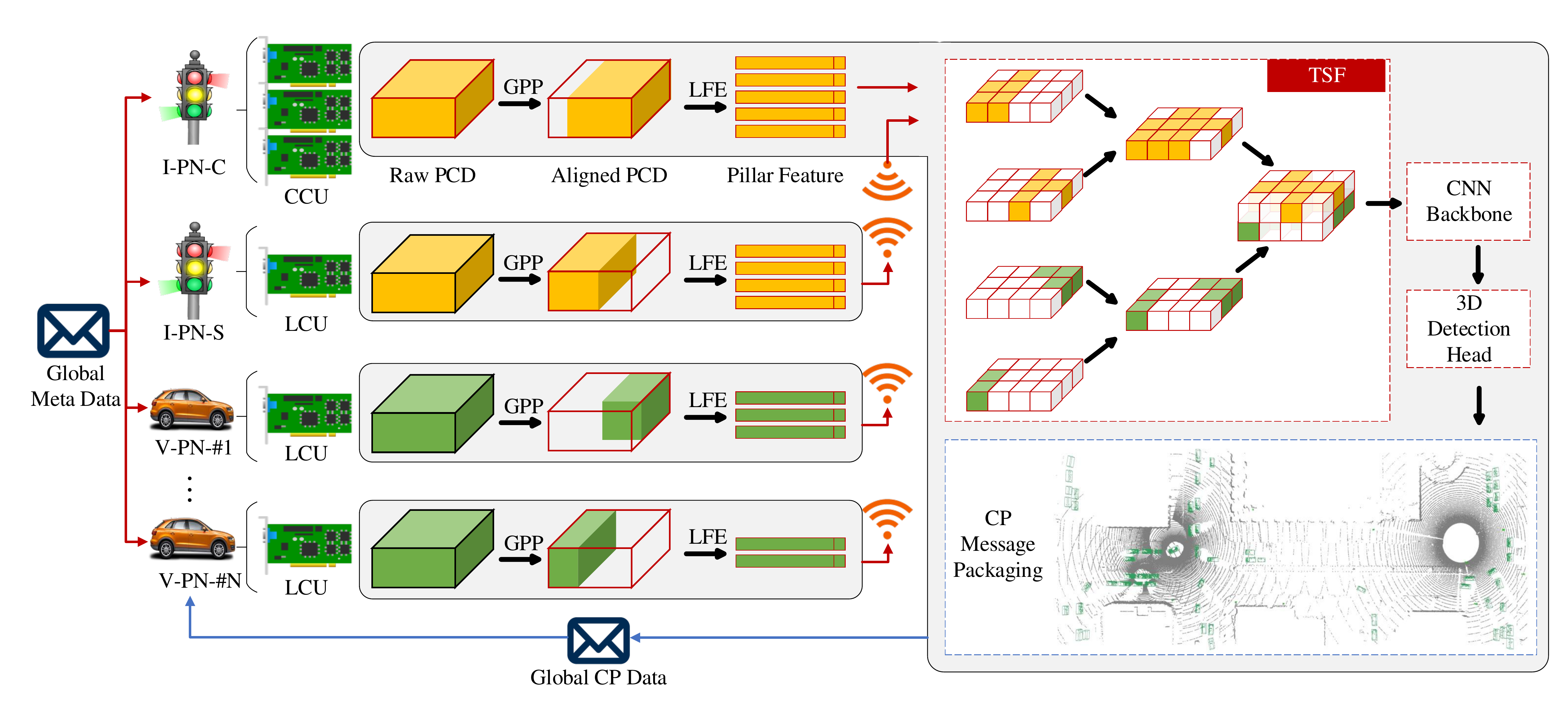}
    \caption{\textbf{Systematic diagram of the proposed cooperative perception system -- VINet.} It consists of five components:  Global Pre-Processing (GPP), Lightweight Feature Extraction (LFE), Two-Stream Fusion (TSF), Central Feature Backbone (CFB), and 3D Detection Head (3DH).}
    \label{fig:vinet-overview}
\end{figure*}

\subsection{Overview of VINet}
The main purpose of VINet is to reduce the system-wide computational cost by dividing the whole perception pipeline into two phases: 1) the lightweight feature extraction, and 2) the feature fusing and processing. Additionally, perception nodes (PNs) are categorized into two types: 1) \textit{Central Nodes} which are equipped with powerful computing devices, named the Central Computing Unit (CCU), and 2) \textit{Slave Nodes} which are equipped with Light-weight Computing Unit (LCU). In this paper, illustrated in Figure~\ref{fig:vinet-overview}, one intersection PN (I-PN) is assigned as Central Node (I-PN-C) while other vehicle-PNs (V-PNs) and I-PN-S (I-PN-Slave) are assigned as Slave Nodes. Therefore, by assigning a lightweight feature extraction pipeline to Slave Nodes, only LCUs are needed for V-PNs and the I-PN-S. Thus, the whole cooperative perception (CP) system will only need one CCU and several LCUs to make CP available for every node with connectivity.

The network overview of VINet is shown in Figure~\ref{fig:vinet-overview}. To reduce inter-node communication costs, Global Meta Data is designed and transmitted to all the PNs. From the system perspective, unlike previous CP methods which require inter-node local-coordinate transformation~\cite{wang2020v2vnet,xu2022v2x} with a communication cost of $\mathbf{O}(N^{2})$, VINet only needs global transformation once for each node which has the communication cost of $\mathbf{O}(N)$ ($N$ represents the number of PNs).

Based on Figure~\ref{fig:vinet-overview}, the workflow of VINet can be summarized as follows:
\begin{itemize}
    \item Global Pre-Processing (GPP): Based on LCU, raw PCD are transformed to the global coordinate and then geo-fenced within the predefined CP range.
    \item Lightweight Feature Extraction (LFE): based on LCU, globally aligned PCD are fed into a pillar feature extraction network to generate hidden features and transmit them to the Central Node.
    \item Two-Stream Fusion (TSF): based on CCU, pillar features from heterogeneous nodes are fused into a unified feature map.
    \item Central Feature Backbone (CFB): based on CCU, a multi-scale CNN backbone generate a feature map from fused feature data.
    \item 3D Detection Head (3DH): based on CCU, detected bounding boxes are generated with class types for further broadcasting to all the connected nodes.
\end{itemize}

The following sections will introduce the aforementioned modules and details about the loss function applied in VINet.

\subsection{Global Pre-Processing}

\subsubsection{Global Transformation}
    
For global coordinate referencing, we design sensor calibration matrices that include the global 3D location and rotation information of the sensor, which is named as \textit{Sensor Location and Pose} (SLaP) data. Raw PCD is originally recorded and organized in the 3D Cartesian coordinate with the center of the sensor of each node. Currently, one popular way of cooperative data transformation is to transform the PCD to the ego-vehicle's coordinate~\cite{xu2022v2x}, otherwise, there will exist spatial matching issues~\cite{F-cooper}.

Since the PCD is collected based on a 3D Cartesian coordinate centered with the SLaP of the sensor, cooperative transformation is designed to unify the PCD from different sensors. The raw PCD can be described by:
\begin{equation}
    {P} = \{[x, y, z, i]^{T} \, |\, [x, y, z]^{T} \in \mathbf{R}^{3}, i\in [0.0, 1.0] \}
\end{equation}
    
The SLaP information is defined by the 3D location and rotation of the sensor:
\begin{equation}
    \text{SLaP} = [X, Y, Z, P, \Theta, R]
\end{equation}
where $X$, $Y$, $Z$, $P$, $\Theta$, and $R$ represent the 3D location along $x-$axis, $y-$axis, $z-$axis; and the pitch, yaw, and roll angles of the sensors in the global coordinate, respectively.

To avoid square-level inter-node communication which requires unreasonable system-wide communication costs, GPP aims to transform the PCD from PNs' coordinates to a unified static global coordinate. Two main advantages of GPP compared with traditional inter-node transformation can be identified: 1) only linearly increasing communication cost is required for the whole CP system, and 2) a static global coordinate can have much less positioning error compared with the ego-vehicle coordinate.

\subsubsection{Cooperative Geo-fencing}
After the global transformation, a geo-fencing process is applied to PCD. In this paper, the detected region $\Omega$ for each of the LiDAR sensors is defined as a $280m \times 90m$ area centered at the location of the respective LiDAR. Specifically, ${P}^{S\rightarrow G}$ is geo-fenced by:
\begin{equation}
{P}^{S\rightarrow G}_{\Omega} = \{[x, y, z, i]^{T} \, |\, x \in {X}, y \in {Y}, z \in {Z}\}   
\end{equation}
where $P^{S\rightarrow G}_{\Omega}$ represents the 3D point cloud data after geo-fencing; and ${X}$, ${Z}$ and ${Y}$ are defined by the CP range, which is further described in Section~\ref{exp-setup}.

\subsection{Lightweight Feature Extraction}
    
% \begin{figure}[!ht]
%     % \centering
%     \includegraphics[width=\textwidth]{pillarEncoder.png}
%     \caption{The LFE process of extracting the feature from points in pillars.}
%     \label{fig: LFE}
% \end{figure}

Although deep features via the CNN backbone are intuitively suitable for CP tasks~\cite{F-cooper}, it requires that every PNs have the capability of supporting the whole feature extraction pipeline. To distinguish our lightweight feature, the features generated after the whole CNN backbone are named \textit{dense features}. In one of the popular deep fusion-based CP methods, \textit{F-Cooper}~\cite{F-cooper}, dense features need to be calculated for all the involved PNs and then shared with each other. Since a CNN backbone is generally a deep neural network with a large amount of computing requirement~\cite{yan2018second, ren2016faster}, deploying every PNs with powerful computing devices is a significant load for real-world implementation and commercialization. Additionally, dense features usually have a tremendous data size and thus they require powerful data compression techniques~\cite{wang2020v2vnet} to make the pipeline feasible in terms of real-time performance, which may inevitably cause perception performance drop and/or additional computing costs.

To reduce the computing requirement from the perspective of sharing feature extraction, a lightweight feature extraction (LFE) module is designed in this study. Inspired by \cite{lang2019pointpillars}, LFE aims to generate pillar features without the involvement of the CNN backbone, which can significantly reduce the computational loads on PNs. To cope with the heterogeneity of PCD from vehicles and infrastructures, two decoupled LFE blocks, i.e., vehicle LFE (V-LFE) and infrastructure LFE (I-LFE), are designed, which have the same network structure but different parameters. 

For each LFE, the first part is to generate a $D-$dimensional feature vector ${V}_{p}$ for all points in pillars:
\begin{equation}
{V}_{p} =\{\left[x, y, z, i, x_{c}, y_{c}, z_{c}, x_{p}, y_{p} \right]_{i}\}^{N}_{i=1}, p = 1, \dots, P
\end{equation}
where $x_{c}, y_{c}, z_{c}, x_{p}, y_{p}, N$, and $P$ represent the distance of each point to the arithmetic center of all points in the $p-$th pillar (the $c$ subscript) and the geometrical center of the pillar (the $p$ subscript), the number of points in the pillar and the number of pillars, respectively. 

The second part is to extract the deep features of ${V}_{p}$ by the following processes. 
\begin{equation}
    \mathcal{H}_{inf}^{(j)} = \max_{1 \leq i \leq N} (Dense_{inf}(\mathcal{F}_{inf}^{(j)})), j = 1, ..., n
\end{equation}

\begin{equation}
    \mathcal{H}_{veh}^{(j)} = \max_{1 \leq i \leq N} (Dense_{veh}(\mathcal{F}_{veh}^{(j)})), j = 1, ..., m
\end{equation}
\begin{equation}
    \mathcal{F}_{inf}^{(j)} = \{\{{V}_{p}^{(i)}\}^{P_{j}}_{i=1}\}_{j=1}^{n}
\end{equation}
\begin{equation}
    \mathcal{F}_{veh}^{(j)} = \{\{{V}_{p}^{(i)}\}^{P_{j}}_{i=1}\}_{j=1}^{m}
\end{equation}
where $\mathcal{H}_{inf}^{(j)}$ and $\mathcal{H}_{veh}^{(j)}$ present the feature output from the LFE module which have the shape of $(P_{j}, C)$. Two decoupled MLP networks are designed as $Dense_{inf}(\cdot)$ and $Dense_{veh}(\cdot)$ to extract $V_{p}$ from $D-$dimensional to $C-$dimensional. $\mathcal{F}_{inf}^{(j)}$ and $\mathcal{F}_{veh}^{(j)}$ present the feature data from $j-$th infrastructure/vehicle node.

\subsection{Two-Stream Fusion}
\begin{figure}[!ht]
    \centering
    \includegraphics[width=\textwidth]{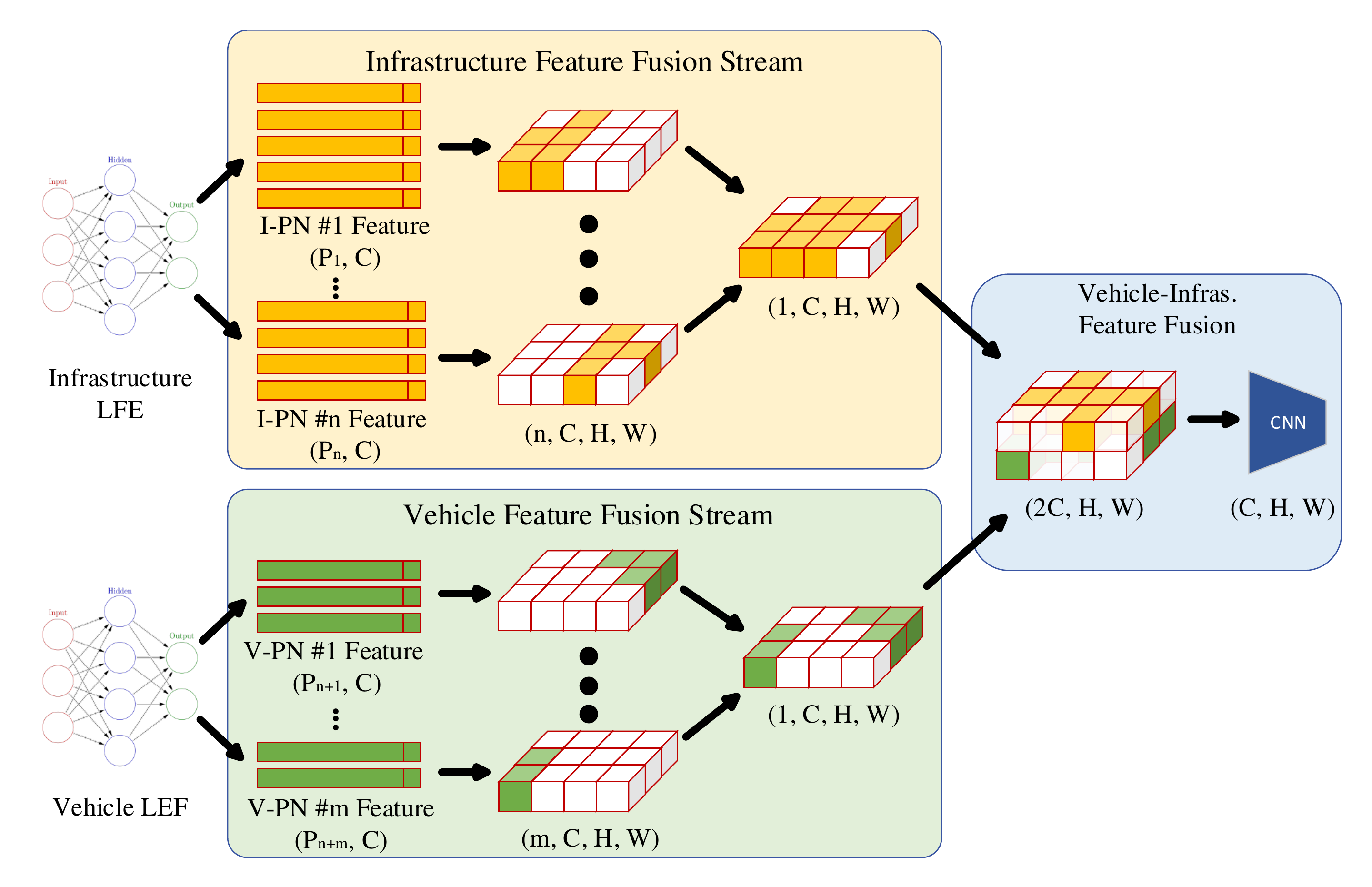}
    \caption{Illustration of the TSF process.}
    \label{fig:VINet-TSF}
\end{figure}

In this section, to combine the features from scalable and heterogeneous PNs, a specific feature fusion model, called \textit{TSF}, is proposed. Since the roadside sensors and onboard sensors have different locations and poses, the point cloud distributions vary based on different sensor configurations. Although all those data can be transformed into a unified coordinate, the differences in data distribution still affect the fusion performance~\cite{bai2022pillargrid, xu2022v2x}. Specifically, TSF consists of two main contributors, which are named \textit{VI (\textbf{V}ehicle-\textbf{I}nfrastructure) Stream} and \textit{VI Fusion}, respectively. The whole fusion process can be formulated as:

\begin{equation}
    \mathcal{S}_{inf} = \max_{1 \leq j \leq n} (Concat(\{\mathcal{M}(\mathcal{H}_{inf}^{(j)})\}_{j = 1}^{n}))
\end{equation}
\begin{equation}
    \mathcal{S}_{inf} = \max_{1 \leq j \leq m} (Concat(\{\mathcal{M}(\mathcal{H}_{veh}^{(j)} )\}_{j = 1}^{m}))
\end{equation}
\begin{equation}
    \mathcal{H}_{cp} = \mathcal{C}(Concat(\mathcal{S}_{inf}, \mathcal{S}_{inf}))
\end{equation}
where $\mathcal{M}(\cdot)$ is the designed mapping function to generate a 2D-pseudo-feature map from the LFE inputs; $Concat(\cdot)$ is the concatenation function to aggregate feature map $H$ in each stream; $\mathcal{S}_{inf}$ and $\mathcal{S}_{inf}$ are the fused feature from each VI stream with a shape of $(1, C, H, W)$; $\mathcal{H}_{cp}$ represents the output of TSF after a concatenation layer $Concat(\cdot)$ and a convolution layer $\mathcal{C}(\cdot)$.

Figure~\ref{fig:VINet-TSF} is provided to further explain the TSF process. To consider the heterogeneity of features from different kinds of PNs, the core ideology of TSF is to create two separate feature streams to decouple the feature extraction and fusion within each class of PNs. For instance, as shown in , two dedicated LFEs are designed for extracting pillar features based on the data from vehicles and infrastructures, respectively. A regrouping process is applied to grouping those pillar features into two groups. 

A feature mapping process is implemented for each group of features to generate a 2D spatial feature map for each PN in this group, e.g., feature tensor with the shape of $(n, C, H, W)$ for the infrastructure stream. Based on the previous GPP process, the feature maps from different PNs are spatially aligned with the global coordinate already. Then, a fusion process is designed to fuse features from different nodes while keeping their heterogeneity. At the end of the VI stream, features from each stream will be extracted into a $(1, C, H, W)$ tensor. The fused features for each stream are then fused using VI fusion to form a $(2C, H, W)$ feature tensor. Finally, a CNN fuses the $(2C, H, W)$ tensor into the shape of $(C, H, W)$ to keep a consistent spatial shape of output.

\subsection{Central Feature Backbone}
Based on the centralized design, VINet only needs one CNN backbone, named Central Feature Backbone (CFB), for the whole CP system for all PNs. To make a fair comparison with others, a classic Region Proposal Network (RPN)~\cite{zhou2018voxelnet} is implemented as the CFB in VINet. It is noted that from a theoretical standpoint, since the central PN is equipped with CCU, CFB can be a more powerful backbone to further squeeze the performance of fused features.

\begin{figure}[!h]
    % \centering
    \includegraphics[width=\textwidth]{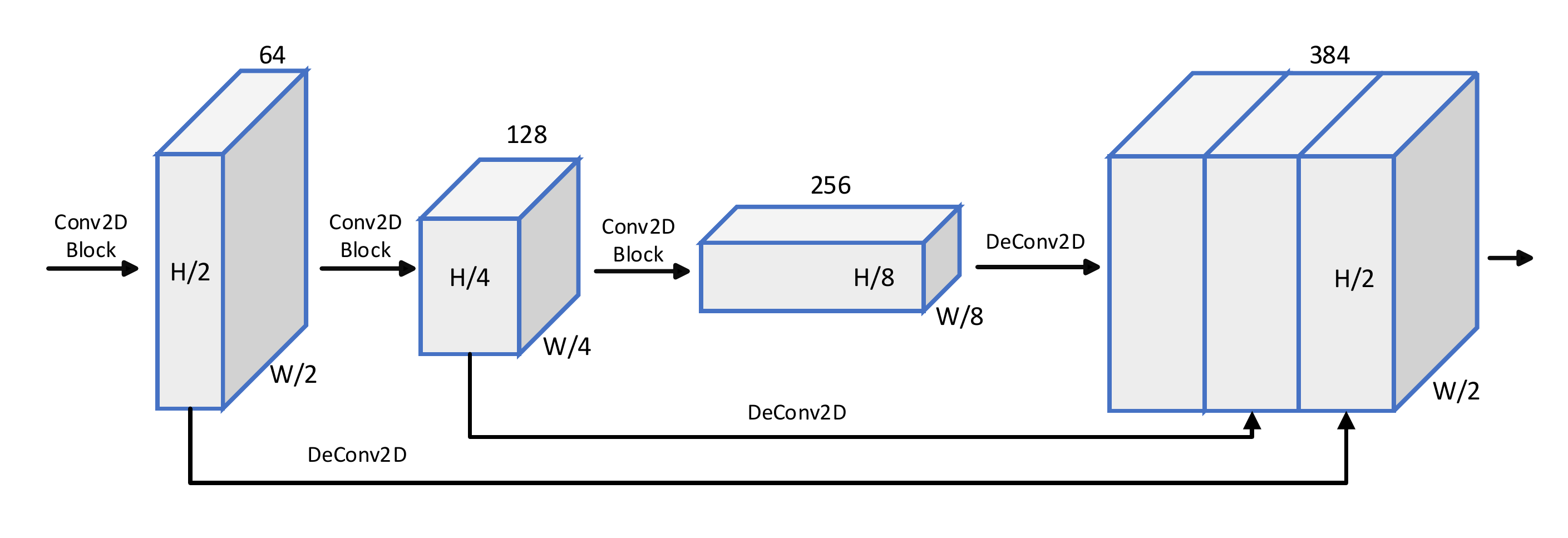}
    \caption{Illustration of the CFB structure which consists of Convolutional and Deconvolutional blocks.}
    \label{fig:rpn_backbone}
\end{figure}
The CFB consists of three phases: 1) \textit{Downsample CNN}, which consists of several 2D CNN blocks (\textit{Conv2D}) to extract the features in a spatial downsampling manner, 2) \textit{Upsample CNN}, which consists of several deconvolutional blocks (\textit{DeConv2D}) to extract the features in a spatially upsampling manner, and 3) \textit{Multi-scale fusion}, which concatenates the output of each \textit{DeConv2D} to gain multi-scale feature for each spatial location. 

\subsection{3D Detection Head}
To generate 3D rotated bounding boxes, an anchor-based 3D dense head~\cite{lang2019pointpillars} is applied for all the models. Intersection over Union (IoU) is implemented to match the prior bounding boxes with the ground truth. Additionally, both pedestrians and vehicles are considered.

\section{Experiments}
\label{sec: experiments}
\subsection{Dataset Acquisition}
\label{Dataset Acquisition}
% \begin{figure}[!h]
%     % \centering
%     \includegraphics[width=0.6\textwidth]{Carla_kitti_transform.pdf}
%     \caption{Illustration of data transformation from the CARLA environment to KITTI's format.}
%     \label{fig:Carla_kitti_transform}
% \end{figure}

Considering the requirement of cooperative perception (CP) research, a platform is needed for customized dataset acquisition. Thus, a CP platform has been designed and developed to support CP model training and validation, such as sensor data collecting and ground truth labeling. We build a CP platform on top of the CARLA simulator~\cite{dosovitskiy2017carla} for data collection. In addition, to ease the re-usability and accessibility, we organize the new dataset to be compatible with existing open-source training platforms (e.g., OpenMMLab~\cite{mmdet3d2020}), by following the KITTI's format~\cite{Geiger2012CVPR}. Nonetheless, the KITTI's settings and the CARLA have different coordinate systems, so a conversion is used to generate the ground truth labels in the KITTI coordinates.

All the sensors are synchronized based on CARLA and the simulation is running at $10$Hz while the data frame is collected at $2$Hz to increase the variety of the whole dataset (i.e., same strategy as NuScenes~\cite{caesar2020nuscenes}). Literally, from our CP platform, unlimited frames of data can be collected with nearly zero cost. In this paper, for speeding up the experimental process, $8,695$ frames of 3D point clouds are collected, including $4347$ frames for training, $1449$ frames for evaluation, and $2899$ frames for testing.

Furthermore, the LiDAR settings for onboard and roadside are different. Detailed specifications of those Lidar sensors are described in Table~\ref{tab:parameter}. The main differences lie in the heights and FOVs.
% Table generated by Excel2LaTeX from sheet 'Sheet5'
\begin{table*}[htbp]
  \centering
  \caption{Parameter Configuration and Description for the 3D-LiDAR Sensors (onboard/roadside) Deployed in the CARTI Platform}
  \resizebox{\textwidth}{!}{%
    \begin{tabular}{c|c|l}
    \toprule
    Parameters  & Default  & \multicolumn{1}{c}{Description} \\
    \midrule
    Channels  & $64$ & Number of lasers. \\
    Height  & $1.74/4.74m $ & Height above the ground. \\
    Range  & $100.0m $ & Maximum distance to measure/ray-cast in meters. \\
    Frequency  & $10.0$Hz & LiDAR rotation frequency. \\
    Upper FOV  & $+22.5~/~+0$ & Angle in degrees of the highest laser beam. \\
    Lower FOV  & $-22.5~/~-22.5$ & Angle in degrees of the lowest laser beam. \\
    Reflection rate  & $0.004$ & Coefficient that measures the LiDAR intensity loss. \\
    Noise stdev  & $0.01$ & The standard deviation of the noise model of points. \\
    Dropoff rate & $45\%$ & General proportion of points that are randomly dropped. \\
    Dropoff intensity & $0.8$ & The threshold of intensity value for exempting dropoff. \\
    Dropoff zero intensity  & $40\%$ &  The probability value of dropoff for zero-intensity points. \\
    \bottomrule
    \end{tabular}}%
  \label{tab:parameter}%
\end{table*}%

\subsection{Experimental Setup}
\label{exp-setup}
For the purpose of evaluating and investigating the CP performance of leveraging multiple PNs, the CP scenario (shown in Figure~\ref{fig: concept}) is set as a two-adjacent intersection area with $280m\times80m$, which is much larger than a single-node perception area. Specifically, in Figure~\ref{fig: concept}, the western infrastructure is defined as the I-PN-C whose coordinate is also set as the reference of the global coordinate. Thus, for GPP, the specific perception field is in the range of $x \in [-53.76m, 181.76m]$ and $y \in [-48.6m, 41m]$. For LFE, both PCD from the roadside LiDARs and onboard LiDARs are transformed into pillars with the voxel size of [$0.23m, 0.23m, 4.0m$] in this paper, the maximum number of pillars for each node is set as $[25,000, 15,000]$ for training and testing. For CFB, each Conv2D block consists of one Conv2D layer with the kernel of $(3, 2, 1)$, followed by several Conv2D layers with kernels of $(3, 1, 1)$. Specifically, the numbers of Conv2D layers in each block are $4$, $6$, and $6$, respectively. For 3DH, the number of object classes is set to be $2$, i.e., car and pedestrian.

\subsubsection{Training Details}
The training and testing platform is equipped with Intel$^\text{\textregistered}$ Core™ i7-10700K CPU@3.80GHz$\times$16 and NVIDIA GPU@GeForce RTX 3090. The training pipeline is designed with 160 epochs with \textit{Batchsize} of 2. During training, a data sample strategy is applied by filtering the ground target by minimum points (MP) reflected by LiDAR. Specifically, MP is set as $5$ and $10$ for car and pedestrian objects, respectively.

\subsubsection{Evaluation Matrix}
% Specifically, the detection results can be categorized into four classes:
\begin{itemize}
    \item \textbf{True Positive (TP)}: the number of cases predicted as positive by the classifier when they are indeed positive, i.e., a vehicle object is detected as a vehicle.
    
    \item \textbf{False Positive (FP)} = the number of cases predicted as positive by the classifier when they are indeed negative, i.e., a non-vehicle object is detected as a vehicle.
    
    \item \textbf{True Negative (TN)} = the number of cases predicted as negative by the classifier when they are indeed negative, i.e., a non-vehicle object is detected as a non-vehicle object.

    \item \textbf{False Negative (FN)} = the number of cases predicted as negative by the classifier when they are indeed positive, i.e., a vehicle is detected as a non-vehicle object.
\end{itemize}

\textit{Precision} is the ability of the detector to identify only relevant objects, i.e., vehicles and pedestrians in this paper. It is the proportion of correct positive predictions and is given by

\begin{equation}
    Precision = \frac{TP}{TP+FP} = \frac{TP}{\text{\# of all detections}}
\end{equation}

\textit{Recall} is a metric that measures the ability of the detector to find all the relevant cases (that is, all the ground truths). It is the proportion of TP detected among all ground truth (i.e., real vehicles) and is defined as

\begin{equation}
    Recall = \frac{TP}{TP+FN} = \frac{TP}{\text{\# of all ground truth}}
\end{equation}

The detection performance is measured with Average Precision (AP) and Average Recall (AR) at Intersection-over-Union (IoU) thresholds of 0.25 for pedestrians and 0.7 for cars, respectively. Furthermore, based on the MP reflected by the ground target, each evaluation class is further divided into three categories: MP$\geq$10, MP$\geq$5, and MP$\geq$1, respectively, to investigate the performance of CP methods on different difficulty levels.

\subsubsection{Compared Baselines}
Three baselines are selected from different perspectives: 1)  \textbf{Raw PCD fusion} -- \textit{EarlyFusion} based on \textit{PointPillar}~\cite{lang2019pointpillars} is considered as it is one of the most straightforward CP methods; 2) \textbf{Dense CNN-feature fusion} -- \textit{F-Cooper}~\cite{F-cooper} is considered since it provides a classic fusion scheme by fusing the feature map generated by the CNN backbone; and 3) \textbf{Pillar feature fusion} -- our last baseline is PillarGrid~\cite{bai2022pillargrid}, the first CP method that uses pillar features for fusion. To make a fair comparison, all methods are trained based on the same CNN backbone and 3D detection head with the same training epochs. All the other data sampling and augmentation techniques are set the same for all methods.

\subsection{Quantitative Evaluation}

% Table generated by Excel2LaTeX from sheet 'Sheet4'
\begin{table*}[!ht]
  \centering
  \caption{AP Results on the CARTI Dataset Using the BEV Detection Benchmark}
  \resizebox{\textwidth}{!}{%
    \begin{tabular}{c|c|c|c|c|c|c|c|c}
    \toprule
    \multirow{2}[4]{*}{Model} & \multirow{2}[4]{*}{Feature Type} & \multirow{2}[4]{*}{Overall AP} & \multicolumn{3}{c|}{Pedestrains AP@0.25} & \multicolumn{3}{c}{Car AP@0.7} \\
\cmidrule{4-9}      &   &   & MP$\geq$10 & MP$\geq$5 & MP$\geq$1 & MP$\geq$10 & MP$\geq$5 & MP$\geq$1 \\
    \midrule
    Early Fusion & Raw PCD & 35.20  & 7.23  & 9.39  & 10.47  & 61.79  & 61.57  & 60.74  \\
    F-Cooper & Dense CNN & 27.46  & 4.38  & 11.63  & 12.34  & 46.35  & 46.34  & 43.71  \\
    PillarGrid & Pillar Feature & 39.48  & 7.49  & 10.81  & 12.87  & 70.42  & 69.13  & 66.17  \\
    VINet & Pillar Feature & \textbf{41.49} & \textbf{8.21} & \textbf{15.48} & \textbf{18.07} & \textbf{70.83} & \textbf{70.11} & \textbf{66.22} \\
    \bottomrule
    \end{tabular}}%
  \label{tab:bevAP}%
\end{table*}%

% Table generated by Excel2LaTeX from sheet 'Sheet4'
\begin{table*}[!ht]
  \centering
  \caption{AP Results on the CARTI Dataset Using the 3D Detection Benchmark}
  \resizebox{\textwidth}{!}{%
    \begin{tabular}{c|c|c|c|c|c|c|c|c}
    \toprule
    \multirow{2}[4]{*}{Model} & \multirow{2}[4]{*}{Feature Type} & \multirow{2}[4]{*}{Overall AP} & \multicolumn{3}{c|}{Pedestrains AP@0.25} & \multicolumn{3}{c}{Car AP@0.7} \\
\cmidrule{4-9}      &   &   & MP$\geq$10 & MP$\geq$5 & MP$\geq$1 & MP$\geq$10 & MP$\geq$5 & MP$\geq$1 \\
    \midrule
    Early Fusion & Raw PCD & 30.26  & 6.94  & 8.56  & 9.34  & 52.39  & 52.27  & 52.08  \\
    F-Cooper & Dense CNN & 24.37  & 4.28  & 11.42  & 12.15  & 40.49  & 39.91  & 37.96  \\
    PillarGrid & Pillar Feature & 35.90  & 6.58  & 10.09  & 12.16  & 62.28  & 62.21  & 62.06  \\
    VINet & Pillar Feature & \textbf{37.81} & \textbf{7.58} & \textbf{14.83} & \textbf{17.16} & \textbf{62.51} & \textbf{62.44} & \textbf{62.31} \\
    \bottomrule
    \end{tabular}}%
  \label{tab:3dAP}%
\end{table*}%

\subsubsection{Object Detection}
The numerical testing results are illustrated in Table~\ref{tab:bevAP} and \ref{tab:3dAP}. Two evaluation benchmarks are involved: the BEV detection benchmark and the 3D detection benchmark. Table~\ref{tab:bevAP} demonstrates the performance on the BEV benchmark. VINet dominates all the compared methods. Specifically, VINet can improve the overall mAP by $+6.29\%$, $+14.03\%$, and $+2.01\%$ compared with Raw PCD fusion~\cite{lang2019pointpillars}, Dense CNN-feature fusion~\cite{F-cooper}, and pillar-feature fusion~\cite{bai2022pillargrid}. Furthermore, for general pedestrian detection (i.e., $MP\geq1$), VINet significantly improves the mAP by over $+5\%$ for all other methods.

For 3D detection results shown in Table~\ref{tab:3dAP}, VINet can offer the mAP improvements by $+7.55\%$, $+13.44\%$, $+1.91\%$ compared with EarlyFusion, F-Cooper, and PillarGrid respectively. Additionally, for car detection, the pillar feature-based methods can bring tremendous performance improvement, such as $10+\%$ mAP than  early fusion and $20+\%$ mAP than Dense CNN-based fusion.

\begin{figure*}[!ht]
    \centering
    \includegraphics[width=\textwidth]{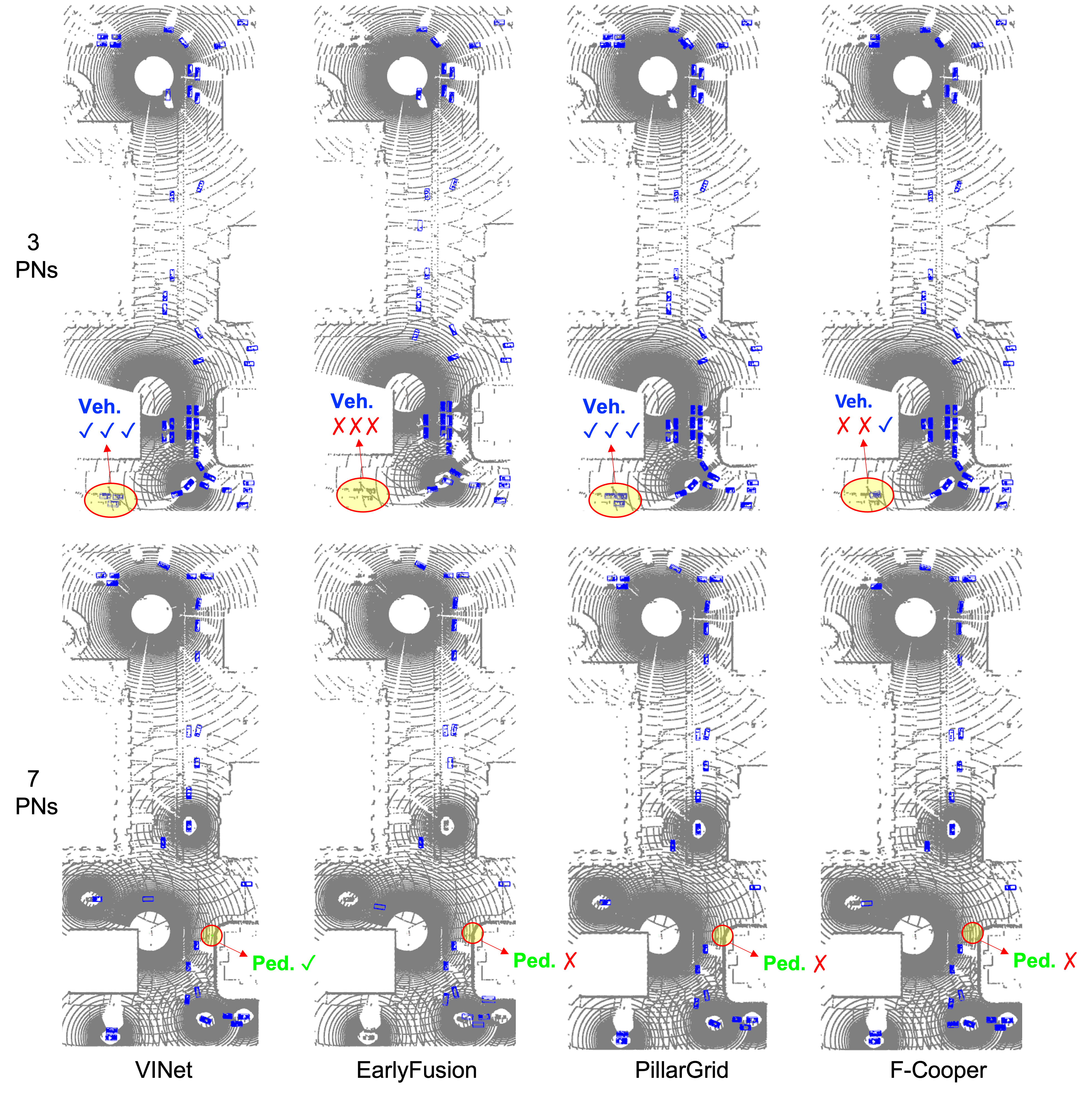}
    \caption{Visualization results for different methods under different numbers of PNs (top row: 3 PNs, bottom row: 7 PNs).}
    \label{fig:vis-results-1}
\end{figure*}

\subsubsection{Real-time Inference}
Table~\ref{tab:cost-eval} also demonstrates the computing speed of the real-time inference pipeline. Both VINet and EarlyFusion can process over 10 frames per second, while F-Cooper can only get 8 frames. Considering that the real-time LiDAR rotating speed is $10Hz$, it demonstrates that VINet also has the capability for real-time inference.

\subsection{Qualitative Results}

Figure~\ref{fig:vis-results-1} shows the visualization CP results of all the methods in the same data frame under different numbers of PNs. 
\subsubsection{Object Detection}
From the perspective of vehicle detection, both VINet and PillarGrid surpass the EarlyFusion and F-Cooper. It is also demonstrated that pillar feature-based cooperation can achieve better performance compared with raw point-based fusion and dense feature-based fusion. In terms of pedestrian detection, VINet surpasses all the baselines, which demonstrates that TSF has a better capability of extracting and fusing features for small objects.  
\subsubsection{Various Number of PNs}
Regarding the number of PNs, when it grows from 3 to 7, feature-based methods, e.g., VINet, PillarGrid, and F-Cooper, can maintain their performance, while EarlyFusion has an evident performance drop. One hypothetical reason is when a single object is perceived by multiple sensors, feature data is more consistent than raw PCD. For instance, if a car is perceived by two LiDARs (one from the front while another from the back), the raw PCD information reflected by this car from those two LiDARs has a bigger difference than the deep feature extracted from the neural networks. In another word, since the neural nets, especially CNN, act like a filtering process, we can infer that feature data from single objects tend to have a smaller divergence than raw PCD. Thus, feature data is also more suitable for large-scale CP systems than raw data.

\subsection{Ablation Study}
In this section, we provide ablation studies and analyze the key design of VINet. From Table~\ref{tab:ablation}, both the VI stream and VI fusion contribute to the detection performance for pedestrians and vehicles. 
% Table generated by Excel2LaTeX from sheet 'Sheet2'
\begin{table}[!ht]
  \centering
  \caption{Component Ablation Study for: i) VI Stream and ii) VI Fusion under 3D Benchmark with MP$\geq$1.}
    \begin{tabular}{c|c|c|c}
    \toprule
    VI Stream & VI Fusion & Pedestrian & Vehicle\\
    \midrule
    $\times$ & $\times$ & 9.73 & 22.15 \\
    \checkmark & $\times$ & 14.33 & 61.41 \\
    $\times$ & \checkmark & 12.16 & 62.07 \\
    \checkmark & \checkmark & \textbf{17.16} & \textbf{62.31} \\
    \bottomrule
    \end{tabular}%
  \label{tab:ablation}%
\end{table}%

\subsubsection{Two Stream Structure}
VI stream significantly impacts the small object detection, i.e., AP $+5\%$ for pedestrians, which implies that the two stream structure can better handle the heterogeneity of sensors.

\subsubsection{Feature Fusion Scheme}
Without the TSF structure, the performance will drop drastically for vehicles, i.e., AP $-40.16\%$, and moderately for pedestrians, i.e. AP $-4.60\%$. This indicates that the TSF plays a vital role for feature fusion. A hypothesis is that vehicles are much larger than pedestrians, and therefore their detection can benefit more from the fused features than only from the single feature.

\section{System-Level Cost Analysis}
In this section, we will comprehensively analyze the computational cost and communication cost from the standpoint of system-level application. We categorize cooperative perception into two types: egocentric CP and holistic CP, which are distinguished by the difference in beneficial nodes. For instance, in most of the recent cooperative perception works~\cite{F-cooper, xu2021opv2v, wang2020v2vnet}, a basic hypothesis under these CP methods is that only one perception node is performing cooperative perception while other nodes only act as the subsidiary to provide feature information to benefit the ego-vehicle. Thus, we define this kind of CP as egocentric CP.

% For instance, in egocentric CP, each vehicle processes the sensor data from others and the processed information would not be shared. Therefore, every vehicle has to perform similar computation tasks, leading to certain computational redundancy. 

Conversely, if all the vehicles are permutation-invariant or, in another word, every vehicle will share information with others to enhance everyone's perception in the network, we define this type of CP as holistic CP. From the perspective of realistic implementation, it may be more beneficial to implement the holistic CP system in the real world. However, the computing complexity of a holistic CP system would be extremely high if no specific optimizations are conducted to the perception structure. 

Thus, we evaluate VINet by comparing it with multiple different baselines as well as under different cooperative perception conditions. Both theoretical analysis and experimental analysis are conducted and shown below. 

\subsection{Theoretical Cost Analysis}

\subsubsection{Computational Complexity}
To analyze the computational complexity of a model, the total number of computations the model has to perform is an important factor. In this paper, we adopt the Floating Point Operations (FLOPs), which apply to any type of computing operations that involve a floating point value. Generally, the more FLOPs the model has to perform, the more complex the model is. In addition to FLOPs, we also involve Multiply-Accumulate Computations (MACs) to illustrate the computational complexity by showing the algebraic expression with respect to the number of perception nodes (PNs).

To calculate the overall FLOPs of a model, convolution layers and fully connected layers are mainly considered, which can be calculated by the following equations:
\begin{equation}
    \mathcal{F}(conv) = 2 \times \mathcal{C} \times \mathcal{S} \times \mathcal{O}
\end{equation}
\begin{equation}
    \mathcal{F}(linear) = 2 \times \mathcal{I} \times \mathcal{O}
\end{equation}
where $\mathcal{C}$, $\mathcal{S}$, $\mathcal{I}$, and $\mathcal{O}$ represent the number of channels, kernel shape, input shape, and output shape, respectively.

\begin{table}[!ht]
\centering
\setlength{\extrarowheight}{0pt}
\addtolength{\extrarowheight}{\aboverulesep}
\addtolength{\extrarowheight}{\belowrulesep}
\setlength{\aboverulesep}{0pt}
\setlength{\belowrulesep}{0pt}
% \captionsetup{labelformat=empty}
\caption{GFLOPs analysis under different cooperative perception conditions.}
\resizebox{\linewidth}{!}{%
\begin{tabular}{c|l|ccccc} 
\toprule
\rowcolor[rgb]{0.753,0.753,0.753} \multicolumn{1}{c!{\vrule width \heavyrulewidth}}{\textbf{Model}} & \multicolumn{1}{c!{\vrule width \heavyrulewidth}}{\textbf{GFLOPs}} & \textbf{VINet(Ours)}          & \textbf{EarlyFusion} & \textbf{LateFusion} & \textbf{F-Cooper} & \textbf{PillarGrid}  \\ 
\toprule
{\cellcolor[rgb]{0.753,0.753,0.753}}                                                                 & No cooperation                                                     & 19.88                   & 0.55                 & 0.55                & 0.55              & 0.55                 \\
{\cellcolor[rgb]{0.753,0.753,0.753}}                                                                 & Egocentric CP w/ 10 PN                                            & 198.80                  & 5.50                 & 5.50                & 5.50              & 5.50                 \\
\multirow{-3}{*}{{\cellcolor[rgb]{0.753,0.753,0.753}}\textbf{Encoder}}                               & Holistic CP w/ 10 PN                                              & 198.80                  & 5.50                 & 5.50                & 5.50              & 5.50                 \\ 
\hline
{\cellcolor[rgb]{0.753,0.753,0.753}}                                                                 & No cooperation                                                     & 270.58                  & 289.91               & 289.91              & 289.91            & 289.91               \\
{\cellcolor[rgb]{0.753,0.753,0.753}}                                                                 & Egocentric CP w/ 10 PN                                            & 270.58                  & 289.91               & 2,899.10            & 2,899.10          & 2,899.10             \\
\multirow{-3}{*}{{\cellcolor[rgb]{0.753,0.753,0.753}}\textbf{Backbone}}                              & Holistic CP w/ 10 PN                                              & 270.58                  & 2,899.10             & 2,899.10            & 2,899.10          & 2,899.10             \\ 
\hline
{\cellcolor[rgb]{0.753,0.753,0.753}}                                                                 & No cooperation                                                     & 4.83                    & 4.83                 & 4.83                & 4.83              & 4.83                 \\
{\cellcolor[rgb]{0.753,0.753,0.753}}                                                                 & Egocentric CP w/ 10 PN                                            & 4.83                    & 4.83                 & 48.30               & 4.83              & 4.83                 \\
\multirow{-3}{*}{{\cellcolor[rgb]{0.753,0.753,0.753}}\textbf{Head}}                                  & Holistic CP w/ 10 PN                                              & 4.83                    & 48.30                & 48.30               & 48.30             & 48.30                \\ 
\hline
{\cellcolor[rgb]{0.753,0.753,0.753}}                                                                 & No cooperation                                                     & 295.29                  & 295.29               & 295.29              & 295.29            & 295.29               \\
{\cellcolor[rgb]{0.753,0.753,0.753}}                                                                 & Egocentric CP w/ 10 PN                                            & 474.21                  & 295.29               & 2,952.90            & 2,909.43          & 2,909.43             \\
\multirow{-3}{*}{{\cellcolor[rgb]{0.753,0.753,0.753}}\textbf{Overall}}                               & Holistic CP w/ 10 PN                                              & \textbf{\uline{474.21}} & 2,952.90             & 2,952.90            & 2,952.90          & 2,952.90             \\
\bottomrule
\end{tabular}
}
\label{tab: GFLOPs complexity}
\end{table}

\begin{table}[!ht]
\centering
\setlength{\extrarowheight}{0pt}
\addtolength{\extrarowheight}{\aboverulesep}
\addtolength{\extrarowheight}{\belowrulesep}
\setlength{\aboverulesep}{0pt}
\setlength{\belowrulesep}{0pt}
% \captionsetup{labelformat=empty}
\caption{Computational complexity under different cooperative perception conditions.}
\resizebox{\linewidth}{!}{%
\begin{tabular}{c|l|ccccc} 
\toprule
\rowcolor[rgb]{0.753,0.753,0.753} \multicolumn{1}{c!{\vrule width \heavyrulewidth}}{Model} & \multicolumn{1}{c!{\vrule width \heavyrulewidth}}{Complexity} & VINet(Ours)                 & EarlyFusion   & LateFusion & F-Cooper & PillarGrid  \\ 
\toprule
{\cellcolor[rgb]{0.753,0.753,0.753}}                                                       & Egocentric Coop. w/ 10 PN                                    & $\mathcal{O}(N)$                  & $\mathcal{O}(1)$ & $\mathcal{O}(N)$       & $\mathcal{O}(N)$     & $\mathcal{O}(N)$        \\
\multirow{-2}{*}{{\cellcolor[rgb]{0.753,0.753,0.753}}Encoder}                              & Holistic Coop. w/ 10 PN                                      & $\mathcal{O}(N)$                  & $\mathcal{O}(N)$          & $\mathcal{O}(N)$       & $\mathcal{O}(N)$     & $\mathcal{O}(N)$        \\ 
\hline
{\cellcolor[rgb]{0.753,0.753,0.753}}                                                       & Egocentric Coop. w/ 10 PN                                    & \textbf{$\mathcal{O}(1)$}         & \textbf{$\mathcal{O}(1)$} & $\mathcal{O}(K)$       & $\mathcal{O}(K)$     & $\mathcal{O}(K)$        \\
\multirow{-2}{*}{{\cellcolor[rgb]{0.753,0.753,0.753}}Backbone}                             & Holistic Coop. w/ 10 PN                                      & \textbf{$\mathcal{O}(1)$}         & $\mathcal{O}(K)$          & $\mathcal{O}(K)$       & $\mathcal{O}(K)$     & $\mathcal{O}(K)$        \\ 
\hline
{\cellcolor[rgb]{0.753,0.753,0.753}}                                                       & Egocentric Coop. w/ 10 PN                                    & \textbf{$\mathcal{O}(1)$}         & \textbf{$\mathcal{O}(1)$} & $O(M)$       & $\mathcal{O}(1)$     & $\mathcal{O}(1)$        \\
\multirow{-2}{*}{{\cellcolor[rgb]{0.753,0.753,0.753}}Head}                                 & Holistic Coop. w/ 10 PN                                      & \textbf{$\mathcal{O}(1)$}         & $O(M)$          & $O(M)$       & $O(M)$     & $O(M)$        \\ 
\hline
{\cellcolor[rgb]{0.753,0.753,0.753}}                                                       & Egocentric Coop. w/ 10 PN                                    & \textbf{$\mathcal{O}(N)$}         & \textbf{$\mathcal{O}(N)$} & $\mathcal{O}(N+K+M)$      & $\mathcal{O}(N+K)$    & $\mathcal{O}(N+K)$       \\
\multirow{-2}{*}{{\cellcolor[rgb]{0.753,0.753,0.753}}Overall}                              & Holistic Coop. w/ 10 PN                                      & \textbf{\uline{$\mathcal{O}(N)$}} & $\mathcal{O}(N+K+M)$         & $\mathcal{O}(N+K+M)$      & $\mathcal{O}(N+K+M)$    & $\mathcal{O}(N+K+M)$      \\
\bottomrule
\end{tabular}
}
\label{tab: theoretical computing complexity}
\end{table}

The computational complexity for different models is calculated and shown in Table~\ref{tab: GFLOPs complexity} and \ref{tab: theoretical computing complexity}. Table~\ref{tab: GFLOPs complexity} shows the GFLOPs (one billion FLOPs) for each model under three different perception conditions: 1) no cooperation, 2) egocentric cooperative perception with 10 perception nodes, and 3) holistic cooperative perception with 10 perception nodes. The results demonstrated that our method can reduce the GFLOPs by a large margin under holistic CP conditions. Even under egocentric CP conditions, in which the EarlyFusion method can be regarded as the lowest boundary of the required GFLOPs, our method performs quite close to that boundary.

In addition, Table~\ref{tab: theoretical computing complexity} demonstrates the computational complexity with theoretically concise notations. Our method can achieve constant complexity for the backbone and head and linear complexity for the encoder. On the other hand, most of the compared methods have a linear complexity for the backbone module, i.e. $\mathcal{O}(K)$, which extremely increases the total GFLOPs of the model.

\subsubsection{Communication Complexity}
To theoretically analyze the communication complexity, megabytes (MB) can be used for evaluating the data size that needed to be transmitted via communication. Since metadata transmission requires much less communication bandwidth than feature data transmission, in this study, the system communication cost is mainly estimated based on feature data transmission. VINet is the first CP method that offers linear communication complexity. 

Therefore, the communication complexity can be calculated as:
\begin{equation}
\begin{aligned}
\mathcal{M}_{vinet}(N) &= N \times m_{vinet} \\
\mathcal{M}_{early}(N) &= N \times (N-1) \times m_{early} \\
\mathcal{M}_{dense}(N) &= N \times (N-1) \times m_{dense} 
\end{aligned}
\label{eq: com-complex}
\end{equation}
where $m_{vinet}$, $m_{early}$, and $m_{dense}$ represent the bandwidth cost per channel for VINet, EarlyFusion, and F-Cooper, respectively. Specifically, $m_{early}$ is based on the statement in \cite{cui2022coopernaut}, while $m_{vinet}$ and $m_{dense}$ are calculated based on their feature size. Thus, the communication complexity for different methods can be summarized in Table~\ref{tab:comm-complexity}. Getting benefit from the global CP structure, VINet is the only method that can achieve linear complexity for communication under holistic CP conditions.

\begin{table}[!ht]
\centering
\caption{Communication complexity under different cooperative perception conditions.}
\resizebox{0.8\linewidth}{!}{%
\begin{tabular}{c|c|c|c|c} 
\toprule
MB & VINet & EarlyFusion & F-Cooper & PillarGrid  \\ 
\midrule
Single Transmission                                      & $\mathcal{O}(1)$  & $\mathcal{O}(1)$        & $\mathcal{O}(1)$     & $\mathcal{O}(1)$        \\
Egocentric Coop. w/ 10 PNs                               & $\mathcal{O}(N)$  & $\mathcal{O}(N)$        & $\mathcal{O}(N)$     & $\mathcal{O}(N)$        \\
Holistic Coop. w/ 10 PNs                                 & \uline{$\mathcal{O}(N)$}  & $\mathcal{O}(N^{2})$        & $\mathcal{O}(N^{2})$     & $\mathcal{O}(N^{2})$        \\
\bottomrule
\end{tabular}
}
\label{tab:comm-complexity}
\end{table}

\subsection{Experimental Cost Analysis}
In this paper, we also conduct real-world experiments to analyze the system-level cost. Since the capacity of GPU memory and communication bandwidth cost are crucial factors for the real-world implementation of the deep network-based method and cooperative perception, respectively, these two factors are analyzed by real-world evaluation and estimation. 

\subsubsection{GPU Cost Estimation}
In this study, we use GPU memory consumption as a surrogate and propose a polynomial model to estimate the system-wide computational costs with limited computational power, in a mathematical manner. 

We can start the analysis from the single-node perception process, which can be divided into three main components in terms of the computing process: 1) the voxelization and pillar encoder, 2) the pillar scatters and CNN backbone and 3) the detection head. We denote the computational costs of these computing processes to be $\mathcal{E}$, $\mathcal{B}$, and $\mathcal{D}$, respectively. Then a polynomial model for single-node perception can be designed as:
\begin{equation}
    \mathcal{E} + \mathcal{B} + \mathcal{D} = \mathcal{C}_{single}
\end{equation}
According to Table~\ref{tab:vinet detail}, the VINet computing process can be divided into three phases: 1) GPP + LEF on LCU, 2) TSF + CFB on CCU, and 3) 3DH on CCU. For CP system with N nodes, the computational cost can be defined as:
\begin{equation}
    N \times \mathcal{E} + \mathcal{B} + \mathcal{D} = \mathcal{C}_{vinet}
\end{equation}
Furthermore, for the dense-feature-based CP methods~\cite{F-cooper}, the fusion happens after the CNN backbone. So their computational cost can be defined as: 
\begin{equation}
    N \times (\mathcal{E} + \mathcal{B}) + \mathcal{D} = \mathcal{C}_{dense}
\end{equation}
Both early fusion and late fusion need $\mathcal{C}_{single}$ for each perception node, thus their computational cost can be defined as:
\begin{equation}
    N \times (\mathcal{E} + \mathcal{B} + \mathcal{D}) = \mathcal{C}_{early/late}
\end{equation}

For the CARTI dataset in this paper, we have 2 to 7 perception nodes collected randomly. Here we assume an average number of PNs, i.e., $N = 4$ to process the testing results for GPU memory cost. During the inference with \textit{batchsize} of 1, $\Tilde{\mathcal{C}}_{early}$, $\Tilde{\mathcal{C}}_{vinet}$ and $\Tilde{\mathcal{C}}_{dense}$ are measured as $12.52$, $5.60$, and $10.95$, respectively (in GB). Therefore, by solving the aforementioned polynomial equations, we will get:
\begin{equation}
\begin{aligned}
\mathcal{E} & = 0.823\\
\mathcal{B} & = 1.783\\
\mathcal{D} & = 0.520
\end{aligned}
\end{equation}
Thus, we can get the polynomial representation of the system-wide cost estimation for the models mentioned above:
\begin{equation}
\begin{aligned}
\mathcal{C}_{single}(N) & = 3.13\\
\mathcal{C}_{vinet}(N) & = 0.82\times N + 2.30\\
\mathcal{C}_{dense}(N) & = 2.61\times N + 0.52 \\
\mathcal{C}_{early/late}(N) & = 3.13\times N
\end{aligned}
\end{equation}

We use the inference test with \textit{Batchsize$=1$} which can be regarded as the way they work in the real world. Since the CARTI dataset contains 2 to 7 PNs, we use the average system-wide cost for the CARTI dataset testing results as shown in Table~\ref{tab:cost-eval}. The estimation at 10 PNs is also listed in Table~\ref{tab:cost-eval}. 

\begin{table}[!ht]
\centering
% \captionsetup{labelformat=empty}
\caption{GPU memory cost analysis under different cooperative perception conditions.}
\resizebox{\linewidth}{!}{%
\begin{tabular}{l|c|c|c|c} 
\toprule
\multicolumn{1}{c|}{GPU Memo. [GB]} & Shallow Feature(Ours) & Early Feature & Dense Feature & Late Feature  \\ 
\midrule
No Cooperation                      & 3.13                   & 3.13          & 3.13          & 3.13          \\
Egocentric Coop. Avg.               & 5.60                   & 3.13          & 10.85         & 12.52         \\
Holistic Coop. Avg. Est.            & \uline{\textbf{5.60}}  & 12.52         & 12.52         & 12.52         \\
Egocentric Coop. w/10PN Est.        & 10.50                  & 3.13          & 26.52         & 31.30         \\
Holistic Coop. w/10PN Est.          & \uline{\textbf{10.50}} & 31.30         & 31.30         & 31.30         \\
\bottomrule
\end{tabular}
}
\label{tab:cost-eval}
\end{table}

In Table~\ref{tab:cost-eval}, we use \textit{Shallow Feature} -- the lightweight feature from the Encoder -- to distinguish our methods with \textit{Dense Feature} -- the normally used deep feature from the Backbone \cite{F-cooper, xu2022v2x}. Under egocentric CP conditions, the GPU memory consumption of shallow feature-based methods is $48.4\%$ and $55.3\%$ lower than the dense feature-based and late feature-based methods on average. If 10 PNs are involved in the egocentric CP system, our method can reduce GPU memory consumption by $60.4\%$ and $66.5\%$ with respect to dense feature-based and late feature-based methods, respectively. Under holistic CP conditions, our method can reduce $66.5\%$ GPU memory consumption compared with all the other methods.

\begin{figure}[!ht]
    \centering
    \includegraphics[width=\textwidth]{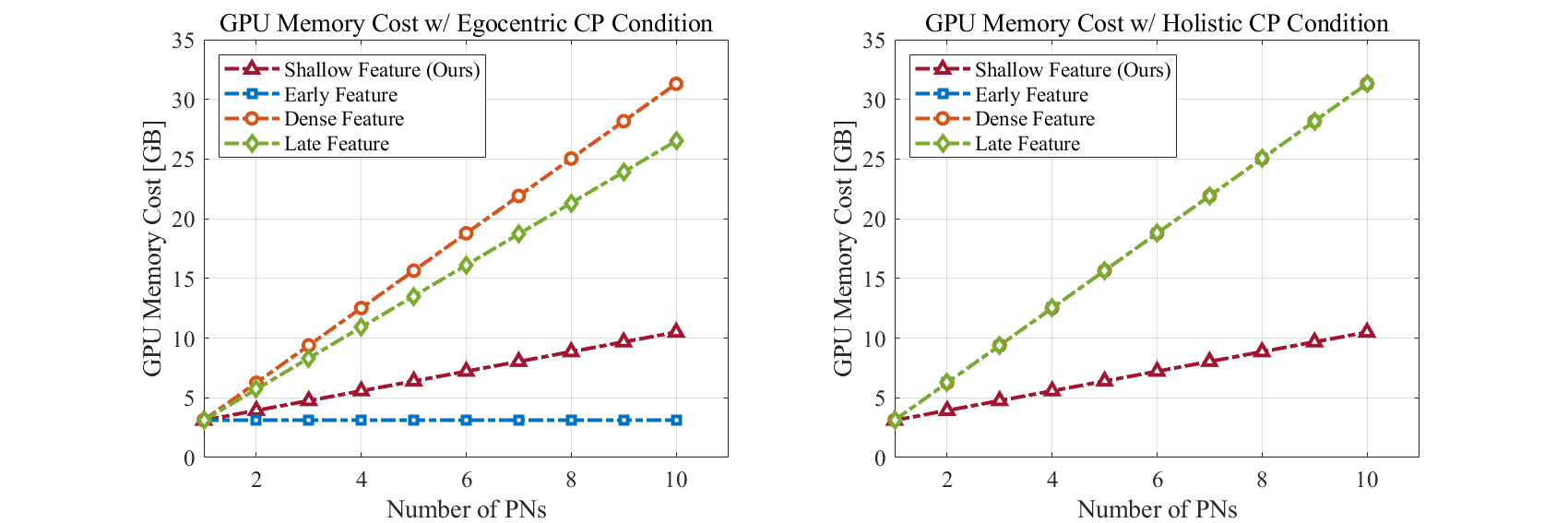}
    \caption{GPU Memory Cost under different CP Condition.}
    \label{fig:gpucost}
\end{figure}

In addition, Figure~\ref{fig:gpucost} demonstrates the change in GPU Memory Cost with different PNs involved in the CP system. It is shown that shallow features are very efficient in terms of GPU memory consumption. Although the early feature requires the lowest GPU memory under egocentric CP conditions, its GPU cost is the same as the late feature and dense feature due to its fundamental system structure.

\subsubsection{Bandwidth Cost Estimation}
For bandwidth cost, we estimate the data size of the feature needed to be transmitted. Specifically, the size for feature per transmission can be roughly formulated using the following equation:
\begin{equation}
    m =  \frac{f \times c \times b}{1,000,000}
\end{equation}
where $f$, $c$, and $b$ represent the number of feature grids, the number of channels in each feature grid, and the number of bytes for each data point (4 for \textit{float32} type used in this paper), respectively. Based on the communication complexity (see Equation~\ref{eq: com-complex}), the bandwidth requirement analysis is shown in Table~\ref{tab:comm-complexity}.

\begin{table}[!ht]
\centering
\caption{Bandwidth requirement analysis under different cooperative perception conditions.}
\resizebox{\linewidth}{!}{%
\begin{tabular}{c|c|c|c|c} 
\toprule
MB & Shallow Feature (Ours) & Early Feature & Dense Feature & Late Feature  \\ 
\midrule
Single Transmission & \uline{3.84}  & 6.00 & \begin{tabular}[c]{@{}c@{}}200 / 6.87\tablefootnote{For dense-feature sharing, data compression is generally necessary to make data transmission possible. Here a $32\times$ compression rate is assumed to F-Cooper for communication cost analysis, based on other dense-feature sharing studies, e.g.,~\cite{xu2022v2x}.}\end{tabular} & 0.025        \\
Egocentric CP w/ 10 PN  &\uline{34.56}  & 54.00       & 61.83   & 0.225       \\
Holistic CP w/ 10 PN   & \uline{34.56} & 540.00      & 618.30  & 2.25      \\
\bottomrule
\end{tabular}
}
\end{table}
\begin{figure}[!ht]
    \centering
    \includegraphics[width=\textwidth]{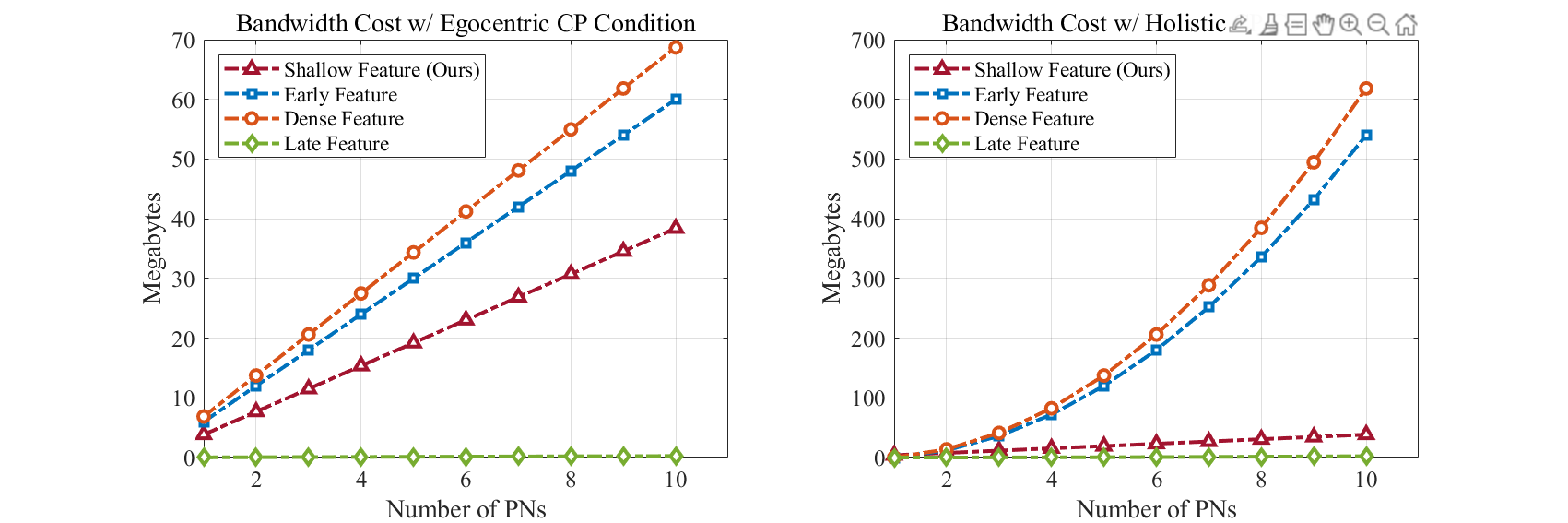}
    \caption{Bandwidth Cost under different CP Condition.}
    \label{fig:commcost}
\end{figure}

From the perspective of feature size, the shallow feature can save $35.7\%$ data space with respect to early features, and 
$44.1\%$ data space with respect to a compressed dense feature. Similar bandwidth reduction can be achieved under egocentric CP conditions. Under holistic CP conditions, our method can reduce $93.6\%$ and $94.4\%$ bandwidth requirements compared with early feature-based and dense feature-based methods.

\section{Conclusions}
\label{sec: conclusion}
In this paper, we propose VINet, the first cooperative object detection method that is designed from the standpoint of large-scale system-level implementation. We demonstrate that for lightweight, scalable, and heterogeneous cooperative 3D object detection tasks, VINet can enhance overall mAP by 1.7\% - 18.9\% for car and pedestrian classes, while reducing system-wide computational costs by 84\% and communication costs by 94\%. Furthermore, VINet also offers a unified framework for cooperative object detection with low system-wide costs. For future work, data compression, impacts of noise levels, and latency will be further investigated to evaluate the performance of VINet under more challenging environments.

\section{Acknowledgments}
This research was funded by Toyota Motor North America, InfoTech Labs. The contents of this paper reflect the views of the authors only, who are responsible for the facts and the accuracy of the data presented herein. The contents do not necessarily reflect the official views of Toyota Motor North America.

%% The Appendices part is started with the command \appendix;
%% appendix sections are then done as normal sections

%% If you have bibdatabase file and want bibtex to generate the
%% bibitems, please use
%%
\bibliographystyle{elsarticle-num} 
\bibliography{vinet}

\newpage
\appendix

\section{Model Details}
In this supplementary material, we first provide more details about the model design and analysis, including the global transformation, loss function, and network configuration. Then, detailed information on the CARTI platform is introduced, followed by the mathematical analysis of system-wide cost estimation. 
\subsection{Global Transformation}
Specifically, the GPP transformation can be defined as:
    
\begin{equation}
    {R}_{X} = \begin{bmatrix}
1 &  0&  0\\
0 &  cos(-R)&  -sin(-R)\\
0 &  sin(-R)&  cos(-R)
\end{bmatrix}
\end{equation}
    
\begin{equation}
    {R}_{Y} = \begin{bmatrix}
cos(-P) &  0&  sin(-P)\\
0 &  1&  0\\
-sin(-P) &  0&  cos(-P)
\end{bmatrix}
\end{equation}

\begin{equation}
    {R}_{Z} = \begin{bmatrix}
cos(-\Theta) &  -sin(-\Theta)&  0\\
sin(-\Theta) &  cos(-\Theta)&  0\\
0 &  0&  1
\end{bmatrix}
\end{equation}

\begin{equation}
{T} = [X\text{\space} Y\text{\space}  Z\text{\space} 0]^T    
\end{equation}

\begin{equation}
{P}^{S\rightarrow G} = 
\begin{bmatrix}
{R}_{X}& 0\\
0&1
\end{bmatrix}
\cdot
\begin{bmatrix}
{R}_{Y}& 0\\
0&1
\end{bmatrix}
\cdot
\begin{bmatrix}
{R}_{Z}& 0\\
0&1
\end{bmatrix}
\cdot{P}^{S} + {T}
\end{equation}
where ${R}_{X}$, ${R}_{Y}$, ${R}_{Z}$, and ${T}$ represent the rotation matrix along $x-$axis, $y-$axis, $z-$axis, and the translation matrix, respectively. ${P}^{S}$ and ${P}^{S\rightarrow G}$ represent the PCD with respect to the sensor's coordinate and global coordinate, respectively.

\subsection{Loss Function}
Regarding a traditional anchor-based detection head, a 7-dimensional vector $(x,y,z,w,l,h,\theta)$ is defined for each box, where $w$, $l$, $h$, and $\theta$ represent the width, length, height, and yaw angle, respectively. The loss function is defined as:
\begin{equation}
    \Delta x = \frac{x^{gt} - x^a}{d^a}, 
    \Delta y = \frac{y^{gt} - y^a}{d^a}, 
    \Delta z = \frac{z^{gt} - z^a}{h^a},
\end{equation}
\begin{equation}  
    \Delta w = \log\frac{w^{gt}}{w^a}, 
    \Delta l = \log\frac{l^{gt}}{l^a}, 
    \Delta h = \log\frac{h^{gt}}{h^a}, 
\end{equation}
\begin{equation}  
    \Delta \theta = sin(\theta^{gt} - \theta^a)
\end{equation}
where the superscript $gt$ and $a$ represent the ground truth and anchor, respectively; and $d^a$ is defined by: 
\begin{equation}
    d^a = \sqrt{(w^a)^2 + (l^a)^2}.
\end{equation}
The total localization loss is:
\begin{equation}
    {L}_{loc} = \sum_{b \in (x, y, z, w, l, h, \theta)} \text{SmoothL1}(\Delta b)
\end{equation}
The object classification loss is defined as:
\begin{equation}
    {L}_{cls} = -\alpha_{a}(1-p^a)^\gamma \log p^a,
\end{equation}
where $p^a$ is the class probability of an anchor; and $\alpha$ and $\gamma$ are set to be $0.25$ and $2$, respectively. Hence, the total loss is:
\begin{equation}
    {L} = \frac{1}{N_{pos}}(\beta_{loc}{L}_{loc} + \beta_{cls}{L}_{cls} + \beta_{dir}{L}_{dir}),
\end{equation}
where $N_{pos}$ is the number of positive anchors; and $\beta_{loc}$, $\beta_{cls}$ and $\beta_{dir}$ are set as $2$, $1$, and $0.2$, respectively.

\subsection{Architectural Specification}
According to the mathematical definitions, the whole VINet model can be formulated as:
\begin{equation}
\begin{aligned}
f_{i} &= GPP(x_{i}),~x_{i} \in \mathcal{R}^{N_{all} \times 4},  &\textit{for $PN_i$, }i = 1, ..., N\\
z_{i} &= LFE(f_{i}),~f_{i} \in {P}^{S\rightarrow G}_{\Omega} , &\textit{for $PN_i$, }i = 1, ..., N\\
z_{f} &= TSF(z_{i}),~z_{i} \in \mathcal{R}^{P \times 64},  &\textit{for Central Node}\\
z_{h} &= CFB(z_{f}),~z_{f} \in \mathcal{R}^{C \times W \times H}, &\textit{for Central Node}\\
y &= 3DH (z_{h}), & \textit{for Central Node}
\end{aligned}
\end{equation}
where $x_{i}$ represents the raw PCD for each PN. Table~\ref{tab:vinet detail} demonstrates the detailed specification of the network structure of VINet.

% Table generated by Excel2LaTeX from sheet 'Sheet3'
\begin{table*}[!ht]
  \centering
  \caption{Detailed Architectural Specifications for VINet}
  \resizebox{\textwidth}{!}{%
    \begin{tabular}{c|c|c|cc}
    \toprule
    Device & Component & Output Size & \multicolumn{2}{c}{VINet Structure} \\
    \midrule
    \multirow{2}[4]{*}{LCU} & GPP & \multirow{2}[4]{*}{N $\times$ [P $\times$ 64]} & \multicolumn{2}{c}{[Voxel Encoder {0.23, 0.23, 4}]} \\
\cmidrule{2-2}\cmidrule{4-5}      & LEF &   & \multicolumn{1}{c|}{[Veh. MLP, 9, 64]} & [Inf. MLP, 9, 64] \\
    \midrule
    \multirow{14}[12]{*}{CCU} & \multirow{4}[8]{*}{TSF} & N $\times$ P $\times$ 64 & \multicolumn{1}{c|}{[Veh. Pillar Scatter (512, 1024)]} & [Inf. Pillar Scatter (512, 1024)] \\
\cmidrule{3-5}      &   & 2 $\times$ [64 $\times$ 512 $\times$ 1024] & \multicolumn{1}{c|}{[Veh. Maxout]} & [Inf. Maxout] \\
\cmidrule{3-5}      &   & 128 $\times$ 512 $\times$ 1024 & \multicolumn{2}{c}{[Concat2, 128]} \\
\cmidrule{3-5}      &   & 64 $\times$ 256 $\times$ 512 & \multicolumn{2}{c}{[Conv3$\times$3, 128, 64, stride 2, BN, ReLU] $\times$ 1} \\
\cmidrule{2-5}      & \multirow{7}[2]{*}{CFB} & \multirow{7}[2]{*}{384 $\times$ 256 $\times$ 512} & \multicolumn{2}{c}{[Conv3$\times$3, 64, 64, stride 1, BN, ReLU] $\times$ 3} \\
      &   &   & \multicolumn{2}{c}{[Conv3$\times$3, 64, 128, stride 2, BN, ReLU] $\times$ 1} \\
      &   &   & \multicolumn{2}{c}{[Conv3$\times$3, 128, 128, stride 1, BN, ReLU] $\times$ 5} \\
      &   &   & \multicolumn{2}{c}{[Conv3$\times$3, 128, 256, stride 2, BN, ReLU] $\times$ 1} \\
      &   &   & \multicolumn{2}{c}{[Conv3$\times$3, 256, 256, stride 1, BN, ReLU] $\times$ 5} \\
      &   &   & \multicolumn{2}{c}{[DeConv3$\times$3, 64, 128, stride 1, BN, ReLU] $\times$ 1} \\
      &   &   & \multicolumn{2}{c}{[DeConv3$\times$3, 128, 128, stride 2, BN, ReLU] $\times$ 1} \\
      &   &   & \multicolumn{2}{c}{[DeConv3$\times$3, 256, 128, stride 4, BN, ReLU] $\times$ 1} \\
      &   &   & \multicolumn{2}{c}{[Concat3, 384]} \\
\cmidrule{2-5}      & \multirow{3}[2]{*}{3DH} & \multirow{3}[2]{*}{256 $\times$ 512 $\times$ 48} & \multicolumn{2}{c}{cls. head: [Conv1$\times$1, 384, 12, stride 1]} \\
      &   &   & \multicolumn{2}{c}{reg. head: [Conv1$\times$1, 384, 28, stride 1]} \\
      &   &   & \multicolumn{2}{c}{dir. head: [Conv1$\times$1, 384, 8, stride 1]} \\
    \bottomrule
    \end{tabular}}%
  \label{tab:vinet detail}%
\end{table*}%

\section{CARTI Platform}
\begin{figure}[!ht]
    \centering
    \includegraphics[width=\textwidth]{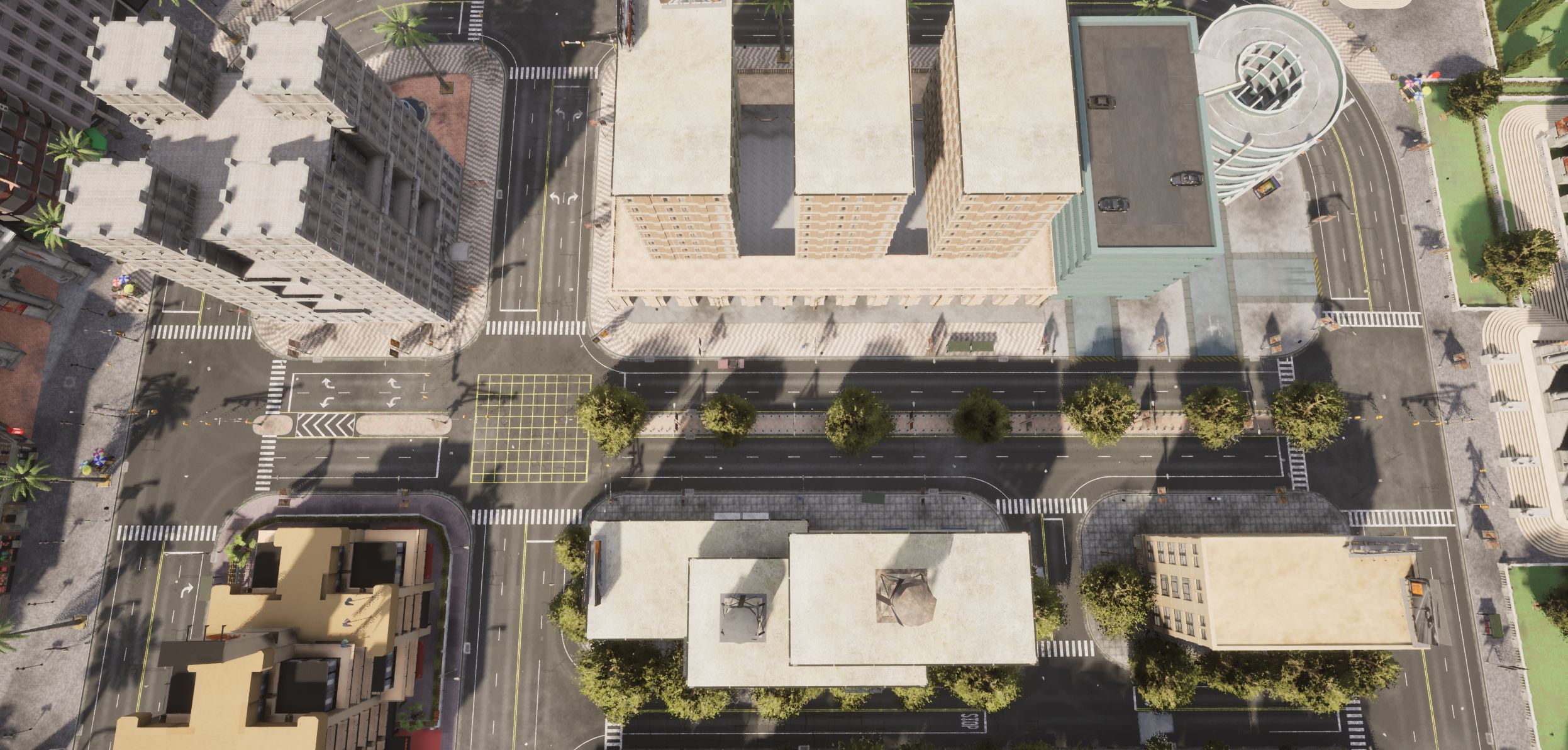}
    \caption{Top-down view for the targeting area of cooperative perception.}
    \label{fig:scene}
\end{figure}

\begin{figure}[!ht]
    \centering
    \includegraphics[width=\textwidth]{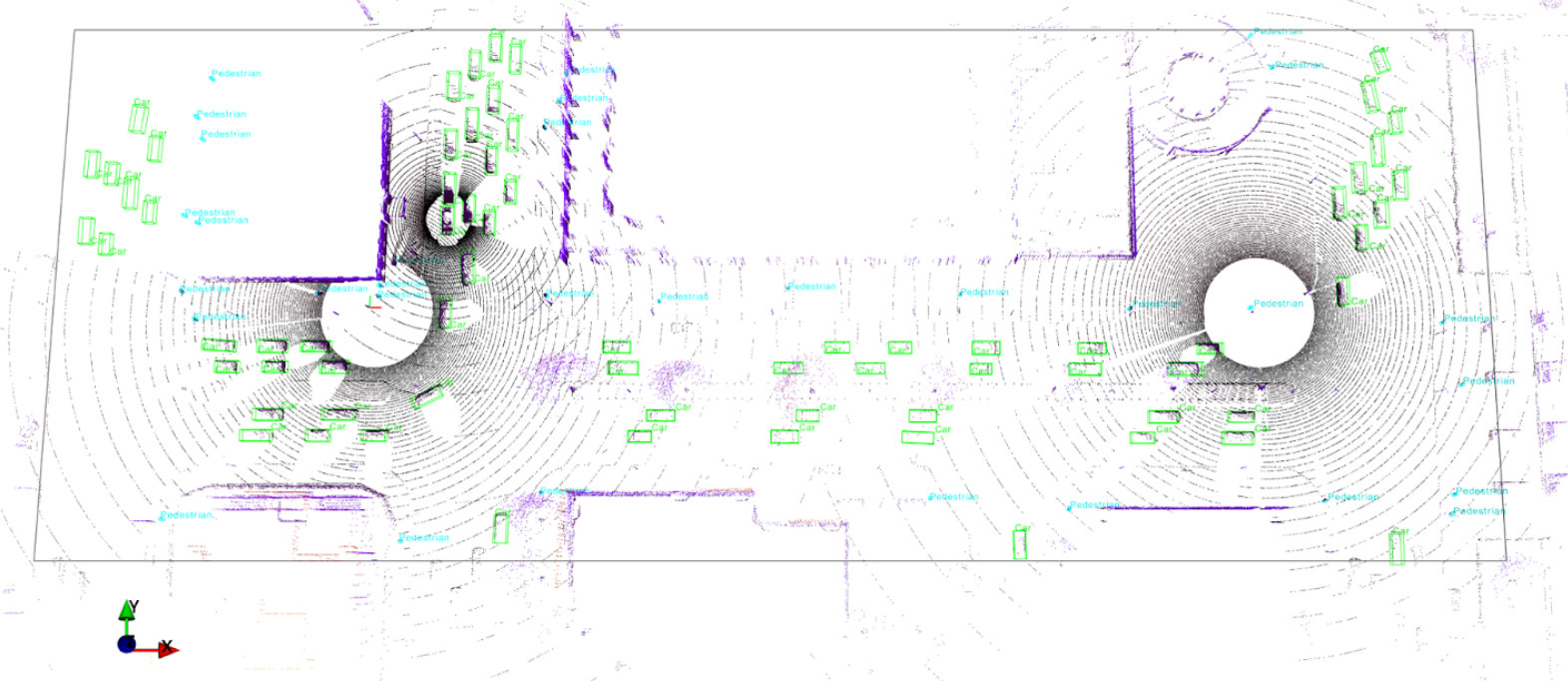}
    \caption{3D point clouds and ground truth labels in one data frame}
    \label{fig:dataset}
\end{figure}

\begin{figure}[!h]
    \centering
    \includegraphics[width=0.8\textwidth]{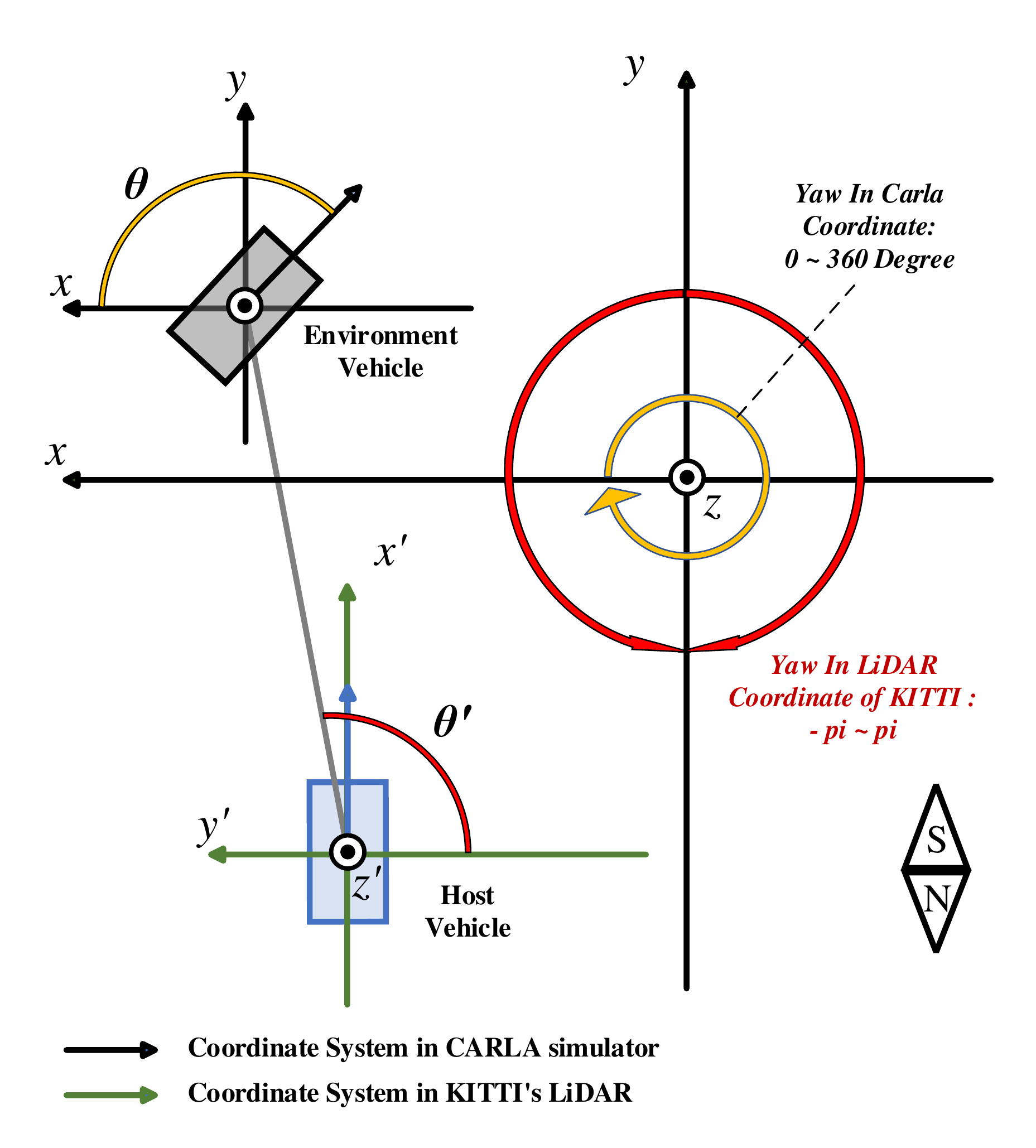}
    \caption{Visualization of data transformation from the CARLA environment to KITTI's format.}
    \label{fig:Carla_kitti_transform}
\end{figure}

The dataset generated from the platform is named ``\textit{CARTI}'' (i.e., \textbf{CAR}la-ki\textbf{T}t\textbf{I}). The CARTI dataset applied in this paper consists of data collected from a varying number of sensors. Specifically, two infrastructure-based LiDAR sensors and $0$ to $5$ vehicle-based LiDAR sensors are deployed in total. Based on our previous experience with the real-world infrastructure-based LiDAR system~\cite{bai2022cmm}, the roadside sensor is equipped on a traffic signal pole of the intersection with a height of $4.74m$. On the other hand, the onboard LiDAR sensors are mounted on top of the CAVs with a height of $1.74m$ based on KITTI's settings. Figure~\ref{fig:Carla_kitti_transform} illustrates the transformation between coordinate systems in CARLA and KITTI.

%% else use the following coding to input the bibitems directly in the
%% TeX file.

% \begin{thebibliography}{00}

% %% \bibitem{label}
% %% Text of bibliographic item

% \bibitem{}

% \end{thebibliography}

\end{document}